\documentclass[10pt,twocolumn,letterpaper]{article}

\usepackage[pagenumbers]{cvpr}      % To produce the REVIEW version
\usepackage[table,xcdraw,usenames,dvipsnames]{xcolor}

\usepackage[utf8]{inputenc} % allow utf-8 input
\usepackage[T1]{fontenc}    % use 8-bit T1 fonts
\usepackage{url}            % simple URL typesetting
\usepackage{booktabs}       % professional-quality tables
\usepackage{amsfonts}       % blackboard math symbols
\usepackage{nicefrac}       % compact symbols for 1/2, etc.
\usepackage{microtype}      % microtypography
\usepackage{adjustbox}
\usepackage{graphicx}
\usepackage{multirow}

\usepackage[linesnumbered,ruled,vlined]{algorithm2e}
\usepackage{amsmath}
\usepackage{amssymb}
\usepackage{bm,bbm}
\usepackage{subcaption}
\usepackage{float}

\newcommand{\ourMethod}[0]{FOMO}
\newcommand{\baseFS}[0]{BASE-FS}
\newcommand{\base}[0]{BASE-ZS+}
\newcommand{\baseLLM}[0]{BASE-ZS+LLM}
\newcommand{\baseIN}[0]{BASE-ZS+IN}
\newcommand{\baseGT}[0]{BASE-ZS+GT}

\usepackage{comment}

\definecolor{cvprblue}{rgb}{0.21,0.49,0.74}
\usepackage[pagebackref,breaklinks,colorlinks,citecolor=cvprblue]{hyperref}

\newcommand{\codebase}[1]{
  \iftoggle{cvprfinal}{
    Our code and benchmark are available at \url{https://orrzohar.github.io/projects/fomo/}.  % If cvprfinal is true
  }{
     Our code and benchmark are available in the supplementary and will be made available upon publication.  
  }
}

\def\paperID{6085} 
\def\confName{CVPR}
\def\confYear{2024}

\title{
Open World Object Detection in the Era of Foundation Models
}
\author{Orr Zohar, Alejandro Lozano, Shelly Goel, Serena Yeung, Kuan-Chieh Wang \\
Stanford University \\
{\tt\small \{orrzohar, lozanoe, shelly23, syyeung, wangkua1\}@stanford.edu} \\
} 

\begin{document}
\maketitle
\begin{abstract}
Object detection is integral to a bevy of real-world applications, from robotics to medical image analysis. To be used reliably in such applications, models must be capable of handling unexpected - or novel - objects. The open world object detection (OWD) paradigm addresses this challenge by enabling models to detect unknown objects and learn discovered ones incrementally. However, OWD method development is hindered due to the stringent benchmark and task definitions. These definitions effectively prohibit foundation models. Here, we aim to relax these definitions and investigate the utilization of pre-trained foundation models in OWD. First, we show that existing benchmarks are insufficient in evaluating methods that utilize foundation models, as even naive integration methods nearly saturate these benchmarks. This result motivated us to curate a new and challenging benchmark for these models. Therefore, we introduce a new benchmark that includes five real-world application-driven datasets, including challenging domains such as aerial and surgical images, and establish baselines. We exploit the inherent connection between classes in application-driven datasets and introduce a novel method, \underline{F}oundation \underline{O}bject detection \underline{M}odel for the \underline{O}pen world, or \ourMethod, which identifies unknown objects based on their shared attributes with the base known objects. \ourMethod\ has $\sim 3\times$  unknown object mAP compared to baselines on our benchmark. However, our results indicate a significant place for improvement - suggesting a great research opportunity in further scaling object detection methods to real-world domains. 
\codebase\
\vspace{-0.15in}

\end{abstract}

\begin{figure}[t]
  \centering
    \resizebox{1\linewidth}{!}{
  \includegraphics{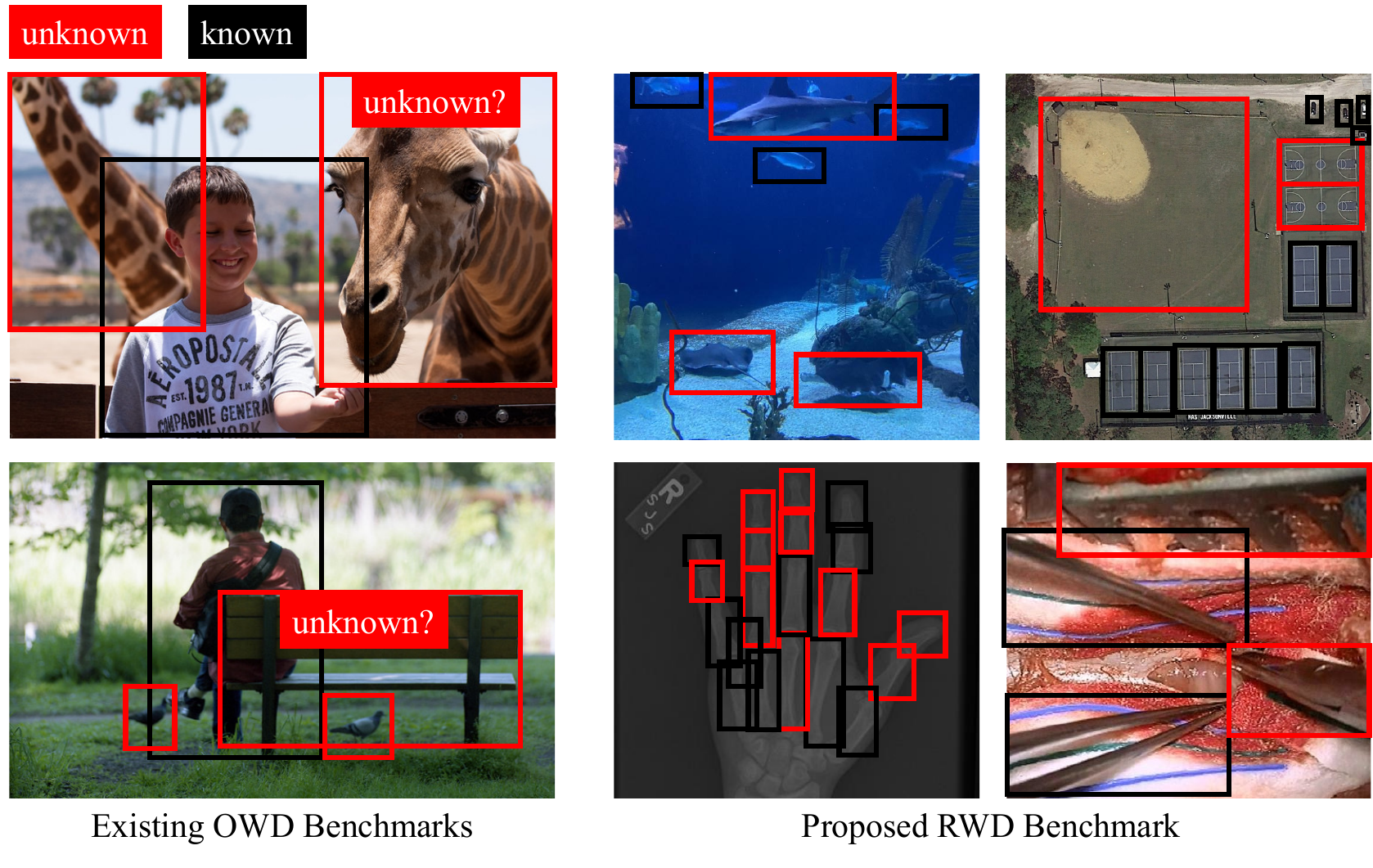}
  }
  \vspace{-0.25in}
  \caption{\textbf{Should `giraffe' and `bench' be \emph{unknown} objects?} Existing open world detection benchmarks are based on the COCO dataset (shown in \emph{left}), which contains arguably the most common objects in everyday environments. 
  This setup also limits the usage of state-of-the-art detection methods based on foundation models, as it is unreasonable that these models did not receive extensive supervision of these objects.
  This work proposes a new application-driven benchmark (shown in \emph{right}), which is out-of-distribution for these foundation models (Sec.~\ref{sec:rwod}). 
  More importantly, this allows us to explore using modern foundation model-based methods.
  }
  \vspace{-0.05in}
\label{fig:poll}
\end{figure}

\section{Introduction }
\label{sec:intro}
Object detection is a foundational computer vision task with applications in a variety of fields such as autonomous driving~\cite{Driving_OWOD}, robotics~\cite{Robots1, Robots2}, and medicine~\cite{HC1, HC2}.
To ensure reliable detection in the face of new environments with \emph{unknown} objects,
open world object detection (OWD) has recently been proposed as a practical computer vision task~\cite{TowardsOWOD}. 
OWD extends detection methods beyond the closed-set assumption, necessitating the detection of both \emph{known/expected} objects and \emph{unknown/novel} objects. 
It also challenges them to incrementally learn previously unknown objects using limited supervision. 
Models that excel in this setting should demonstrate robust detection of unknown objects and flexible learning of those discovered~\cite{TowardsOWOD, OWDETR, CAT, PROB}. 
One would expect the progress in OWD to lead to a broader adaptation of detection methods in applications.

Ensuring that progress in the OWD task truly translates to more reliable real-world detection methods is essential. 
Yet, a discrepancy exists between existing OWD benchmark definitions and the state-of-the-art foundation model-based detection methods. 
On the one hand, the established OWD benchmarks are repurposed from the COCO dataset, containing only 80 everyday classes~\cite{MSCOCO}. Existing OWD benchmarks conceal the class label from a subset of these classes and treat them as `\emph{unknowns}'.  
Yet, should we expect a `bench' to be an unknown object to a foundation detection model (see Fig.~\ref{fig:poll})? 
And, would improvements in detecting such simple classes as unknown indeed translate to making foundation detection methods more reliable? 
On the other hand, these detection methods are trained on large datasets and utilize image-level pre-training at an even larger scale.

Leveraging foundation models in object detection has shown significant promise, with robust performance and capabilities. 
Object detection models that leverage foundation models typically utilize pre-trained vision-language models (VLMs) such as CLIP~\cite{CLIP} before fine-tuning them for detection~\cite{owl-vit, f-vlm}.
As such, it is difficult to determine the degree to which a model was exposed to a particular object - making it impossible to integrate such methods into the original OWD task -
motivating the need for a new benchmark.  
Many foundation models - namely open vocabulary object detection models - extend beyond the closed-set assumption with `text-conditioned' object detection. This approach seeks to recognize objects absent in the training data using their class name via aligned multi-modal embeddings. However, these methods do not attempt to detect novel/unexpected objects.

To enable the reliable utilization of foundation model-based object detection methods in real-world applications, we believe that open world object detection concepts must be integrated. We, therefore, set out to test how these models perform on existing OWD benchmarks. 
We utilized minimal approaches introduced by Maaz et al.~\cite{MAVL}, such as generic prompts to detect unknown objects.
Upon evaluation on OWD benchmarks, we found that even simple foundation model baselines nearly saturate the benchmarks.
As a result, we curated a novel benchmark that combines multiple datasets from diverse real-world applications, including underwater, aerial, and medical domains (see Fig.~\ref{fig:poll}).

Herein, we introduce \textbf{F}oundation \textbf{O}bject detection \textbf{M}odel for the \textbf{O}pen world, or \textbf{\ourMethod}, 
which utilizes foundation object detection models to detect unknown objects. Specifically, \ourMethod\ learns to detect object attributes and the mapping between these attributes to the known objects using a few object exemplars. 
To do this, \ourMethod\ utilizes a few object examples to select and refine attribute embeddings initially proposed by a large language model.
In most practical applications, we found that unknown objects share visual/functional attributes with the base classes~\cite{rosch1975family}. 
Unlike previous OWD methods, we identify unknown objects that are in-distribution to the attributes but out-of-distribution to the known classes. \textbf{Our contributions can be summarized as follows:}

\begin{enumerate}
    \item We first show that existing OWD benchmarks are insufficient when evaluating methods leveraging foundation models - by showing that even naive implementations nearly saturate performance, achieving a U-Recall of $69.0$ and final known mAP of $55.5$.
    \item We curated a novel and challenging application-driven open world object detection benchmark comprised of datasets from real-world applications and domains, including underwater, aerial, and medical domains.   
    \item We introduce 
    \ourMethod\ which leverages pre-trained foundation models to detect unknown objects based on the known class attributes.
    \item  We empirically validated \ourMethod\ on this benchmark compared to baselines, showing an improvement of at least $8\%$ known mAP and $3\times$ in unknown mAP.
\end{enumerate}
 
\section{Related Works}
\label{sec:related}
In this section, we discuss three tasks related to foundation model-based object detection in the open world: open-vocabulary, open world, and class-agnostic object detection.

\paragraph{Open-Vocabulary Object Detection.}  Open-vocabulary object Detection (OVD) is the first field that leveraged foundation models in object detection.
Many methods have emerged following the proposal of the OVD benchmark~\citep{OVD}.  
ViLD~\citep{vild} learns the object detection module by distilling the embeddings on cropped image regions proposed by an off-the-shelf region proposal network. 
RegionCLIP~\citep{zhong2022regionclip} modifies the pre-training stage to account for region information while utilizing pre-trained VLMs. 
In contrast to the above, recent methods like OWL-ViT~\citep{owl-vit, owl-vit-v2} and F-VLM~\citep{f-vlm} by leveraging existing public object detection datasets to learn the detection module, the performance can be substantially improved. 
However, the studies referenced above do not address the capacity of these models to detect relevant unknown objects, as is the case in open world  object detection, as discussed further below. 
To our knowledge, our work is the first to extend OVD methods to the open world.

\paragraph{Open World Object Detection.}
The open world object detection (OWD) task, recently introduced by Joseph et al.~\cite{TowardsOWOD}, has already garnered much attention~\cite{OWDETR, RandBox, UC-OWOD, CAT, PROB, OWOD_OCPL, two_branch_OWOD} due to its possible real-world impact.  Their work introduced ORE, which adapted faster-RCNN with feature-space contrastive clustering, an RPN-based unknown detector, and an Energy-Based Unknown Identifier (EBUI) for the OWD objective. ~\cite{OWOD_OCPL} attempted to extend ORE by minimizing the overlapping distributions of the known and unknown classes in the embeddings featurespace by setting the number of feature clusters to the number of classes, thereby reducing confusion between known and unknown objects. Meanwhile,~\cite{PROB} attempted to extend the deformable DETR model by introducing probabilistic objectness and a modified inference scheme. 
Despite their potential, foundation models have been largely overlooked in OWD, unlike in the OVD paradigm. As research progresses, the propensity for using pre-trained foundation models in object detection is expected to surpass traditional training models from scratch. Our work investigates the use of foundation models within OWD tasks, emphasizing detecting relevant unknown objects.

\paragraph{Class-Agnostic Object Detection.}
In the open world object detection task, a crucial aspect pertains to learning to detect ``unknown'' objects, which mandates the model to acquire a notion of generic objectness.
The class-agnostic object detection task explicitly investigates this ability of object detection models.
In the class-agnostic object detection paradigm, the models aim to detect and localize objects accurately without assigning a categorical label to them. 
A recent approach for the class-agnostic object detection task, MAVL~\citep{MAVL}, builds on multi-modal vision transformers (MViTs) that have been shown to exhibit excellent object detection capability.  
After training on a collection of aligned image-text pairs, the resulting model achieves impressive class-agnostic object detection performance by using intuitive prompts like "all objects".
Inspired by this, our proposed baseline method extends OWL-ViT by using MAVL style prompts for ``unknown'' detection. 
Interestingly,~\cite{MAVL} observed that their model had trouble generalizing to real-world domains such as medical images (their App. B).

\section{Real-World Object Detection Benchmark}
\label{sec:rwod}
The goal of open world object detection is to develop methods that act robustly in the real world - specifically, detecting \emph{unknown} objects and, once classified, incrementally learn them.
Existing OWD benchmarks have attempted to simulate such an environment by re-purposing popular datasets, like COCO, by concealing the class labels for a subset of the classes, thereby introducing \emph{unknown} objects (e.g.,  `\texttt{bench}' as shown in Fig.~\ref{fig:poll}).
While initially slated to be a pragmatic benchmark for developing real-world detection methods, such benchmarks deviate from their original purpose. 
First, they are \emph{unrealistic} as the said `unknowns' are common everyday objects that most modern detection methods can detect well.
Second, the arbitrary split between known and unknown does not emulate real applications. 
In a real detection task, such as detecting animals, unknown classes should be contextually related to the known classes. 
Needlessly detecting many other unrelated objects causes the `unknown' object detection to be irrelevant.
Lastly, a critical issue caused by having an unrealistic benchmark is the requirement to comply with what is considered unknowns, which effectively restricts the incorporation of foundation models.

When we evaluate foundation models on existing OWD benchmarks (see Sec.~\ref{sec:exp:owd}), their performance is effectively saturated. 
However, as we will later show (see Sec.~\ref{sec:exp:rowd}), these same models fail when challenged in real-world application datasets. %\SY{foreshadow that we will indeed see this in your experimental results} 
This inspired us to curate the Real-World Object Detection (RWD) benchmark that challenges OWD methods that leverage foundation models by selecting application-driven datasets with diverse domain and content shifts. In Sec.~\ref{b:dataset_details}, we introduce the different datasets and applications that constitute the benchmark. In Sec.~\ref{b:benchmark_construction}, we detail the construction of the benchmark itself. The proposed evaluation focuses on the \textit{few-shot/low-data} setting, acknowledging that most applications cannot collect datasets on the scale of traditional benchmarks. Finally, in Sec.~\ref{b:unk_obj}, we explain the fundamental difference in the definition of `unknown objects' between our and previous OWD benchmarks.  Unlike benchmarks based on widely used datasets like COCO, the RWD benchmark is application-driven and better reflects real-world scenarios by incorporating open world concepts.
For more, see Sup. Sec.~\ref{sup:sec:benchmark_details}.

\subsection{Dataset Details} 
\label{b:dataset_details}
\begin{table}[t]
\centering
\adjustbox{width=50mm}{
\begin{tabular}{ c|ccc }
\toprule
Name &  \begin{tabular}{@{}c@{}}\# Train \\ Images \end{tabular} & \begin{tabular}{@{}c@{}}\# Test \\ Images \end{tabular} & \# Classes  \\
\toprule
Aquatic  & 318 & 319 & 7 \\
Aerial &  5000 & 5000 & 20 \\
Game  &   788 & 787 & 59 \\
Medical  &  93 & 89& 12 \\
Surgery  & 912  & 917 & 13 \\
\bottomrule

\end{tabular}}
\vspace{-0.05in}
\caption{\textbf{Datasets Details.} Number of classes and images.
See Sup. Sec.~\ref{sup:sec:benchmark_details} for additional details.}
\vspace{-0.16in}
\label{table:dataset_details}
\end{table}
From the diverse array of datasets in Roboflow 100 (RF100)~\citep{roboflow100}, showcased at the CVinW workshop at CVPR 2023 for its real-world applicability, we selectively utilized three distinct datasets: Aquarium, Team Fight Tactics, and X-ray Rheumatology.
The Aquatic (Aquarium) dataset consists of underwater images of different sea creatures, which potentially can be used in underwater applications. The Game (Team Fight Tactics) dataset comprises of game snapshots containing different avatars and is used to evaluate performance on synthetic data. The Medical (X-ray Rheumatology) dataset consists of hand X-rays where different bones are to be detected, with applications in detecting arthritis, fracture, and structural abnormalities in the hands.  
The Aerial dataset was curated from the DIOR dataset~\cite{DIOR2020}, consisting of aerial images of stadiums/storage containers, and can be used for applications in satellite imaging/intelligence. The Surgery dataset was taken from the NeuroSurgicalTools dataset~\citep{NeuroSurgicalToolsDataset}, captured by neurosurgical microscopes, and includes various surgical tools. 
Tab.~\ref{table:dataset_details} has the per-dataset breakdown of the number of classes and dataset size.

\subsection{Benchmark Construction}
\label{b:benchmark_construction}
The RWD benchmark contains five existing application-driven object detection datasets. Each datasets' classes were then divided into two subsets: the 50\% most and least common classes. 
Methods are evaluated in two stages: Task 1 ($\text{T}_1$) and Task 2 ($\text{T}_2$). In $\text{T}_1$, only the 50\% most common classes are considered known, while the 50\% least common classes remain unknown to the models. 
This design choice is well motivated by related open world works, which incorporate it as an extension of the long-tail paradigm~\cite{LongTailOW, LongTailOW_2}.
Models are expected to detect both known and unknown during this stage. 
This evaluation stage tests the ability of object detectors to detect novel/unknown objects. 
In $\text{T}_2$, the remaining 50\% least common classes are revealed, and models are evaluated based on their performance on the set of Previously/Currently known classes.

\subsection{What is an Unknown Object?}
\label{b:unk_obj}
Previous OWD methods have reported only the \textit{unknown recall} (U-Recall), as the notion of what an object is was poorly defined. 
U-Recall is inherently flawed as it does not weigh the accuracy of unknown object prediction. 
Unlike previous works, we have clearly defined objects whose detection is desirable - the objects that were held out for that task.
As RWD only contains real-world application datasets, the classes are inherently related, and unknown objects share some visual or functional attributes with the known base classes. This makes the prediction of \textit{only} the held-out objects more feasible, and therefore, we adopt using mAP.

 \section{Leveraging Foundation Models for Open World Object Detection}
\label{sec:method}
\begin{figure*}[t]
  \centering\includegraphics[width=1\textwidth]{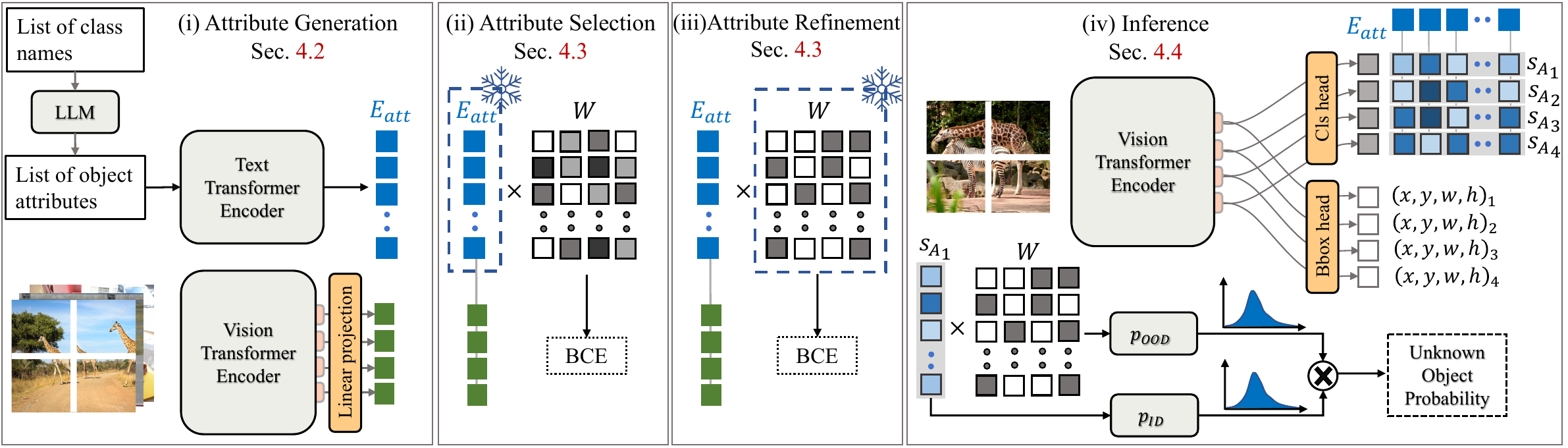}
  \vspace{-0.25in}
  \caption{\textbf{ Overview of \ourMethod.} 
  (i) Attributes are generated using an LLM, which is then encoded using \ourMethod's Text Transformer Encoder into the Attribute Embedding ({\color[HTML]{265FC8}$\mathbf{E}_{att}$}). Meanwhile, vision-based object embeddings are derived from (image-based) object exemplars from the models' Vision Encoder ({\color[HTML]{008000}$\mathbf{e}^v$}).
  (ii) For attribute selection, we update $
  \mathbf{W}$ while freezing $\mathbf{E}_{att}$ using the $BCE$ classification loss, followed by a threshold.
  (iii) to refine the attributes, we update $\mathbf{E}_{att}$ while freezing $
  \mathbf{W}$.
  (iv) An image is fed into the vision encoder during inference, followed by the bounding box and classification heads. The classification head utilizes the pre-computed attribute embeddings to produce the attribute logits. To identify unknown objects, we look for object proposals that are in-distribution (ID) to the attributes but out-of-distribution (OOD) to the known classes. $s_A$ attribute scores between an image and attribute embedding.
  }\vspace{-0.1in}
\label{fig:method}
\end{figure*}

 \textbf{F}oundation \textbf{O}bject detection \textbf{M}odel for the \textbf{O}pen world, or  \ourMethod, directly utilizes foundation models in the open world object detection (OWD) setting (See Fig.~\ref{fig:method}). Essential background is provided in Sec.~\ref{m:background}.
Previous OWD methods never clearly defined an `unknown object', leading to the broad claim that `all' objects are candidate unknown objects. 
This leads to a central question of this paper: ``can we infer what the \emph{unknown} objects are from the given known classes?''  
Inspired by the seminal psychology study by~\citet{rosch1975family} that described \emph{objects} as ``\emph{information-rich bundles of attributes that form natural
discontinuities}'', we reframe detecting unknown objects by reasoning about the attributes of the known classes.

\ourMethod\ attempts to detect unknown objects by identifying objects that share visual and/or functional attributes with the base classes. In Sec.~\ref{m:AttGen}, we present how \ourMethod\ first leverages large language models to identify possible attributes for the target application. As these attributes are class-agnostic, no explicit mapping exists between the attributes and the known object classification. 
Therefore, as described in Sec.~\ref{m:AttSelRef}, \ourMethod\ selects and refines these attributes for the target application by utilizing example images of each class.
In Sec.~\ref{m:inference}, we describe how \ourMethod\ utilizes these attributes to identify known and unknown objects.

\subsection{Background\label{m:background}}
\paragraph{Open World Object Detection (OWD).} 
OWD extends detection methods by removing the closed-set assumption and requiring models to detect and incrementally learn unknown objects. 
During evaluation, we consider multiple stages denoted by $t$. In each stage, the model is given $K^t$ \emph{known classes} of interest denoted using $\mathcal{K}^t = \{C_1^t, C_2^t, ... C_{K^t}^t\}$.
In OWD, the object class label is now a ($K^t+1$)-dimensional vector, where the first element is used to represent the \emph{unknown class}, i.e., $C_0$ denotes the unknown class. 
These unknown objects come from a set of classes of interest whose class names are \emph{not} given to the model, denoted as $\mathcal{U}^t = \{C_{K^t+1}^t, C_{K^t+2}^t, ... C_{K^t+U^t}^t\}$. 
The model then proposes these discovered unknown object classes to an oracle (e.g., a human annotator), which labels the discovered objects. 
The model is then updated to detect the new objects and can now detect $K^{t+1}=K^t+U^t$ object classes. This cycle may be repeated as much as needed.

\paragraph{Foundation Models in Detection.} 
\emph{Foundation models} are models (pre-)trained on large datasets, which demonstrate strong zero-shot and out-of-distribution generalization capabilities (see Bommasani et al.~\cite{bommasani2021opportunities} for an overview on foundation models).
Existing OWD methods have yet to explore the utilization of foundation models.
In contrast, foundation models  have been shown to perform well in open-vocabulary object detection (OVD).

OVD methods undergo two training stages: a first stage focused on learning a multi-modal \textit{representation} using only image-level annotation (e.g., CLIP), and a second stage focused on learning a \textit{detection} model based on the pre-trained multi-modal (language and vision) encoders~\citep{OVD} (more in Sec.~\ref{sec:related}). 
Since the classification is done by matching the text and image embeddings, new classes can be added by simply providing the text prompt with the new class name.
In OVD methods, a \textbf{classification head}  outputs \emph{visual embeddings} denoted as $\mathbf{e}^v \in\mathbb{R}^D$.  Given a class, $C$, specified by the prompt $y_C$ (e.g., ``\texttt{a photo of a <C>}''), the final class prediction is computed between the visual embedding $\mathbf{e}^v$ and text embedding $\mathbf{e}^t_{y_C}$ as follows:
\begin{equation}
    p(C|\mathbf{e}^v,\mathbf{e}^t_{y_C}) = \text{Sigmoid}(\text{Cosine\_Sim}(\mathbf{e}^v,\mathbf{e}^t_{y_C})).
\end{equation} 

OVD methods do not natively support detecting an ``unknown'' object whose class name is not given.  
A naive adaptation for adding this capability is using a generic object prompt like ``\texttt{a photo of an object}'', which we denote as $y_0$, in place of the \emph{unknown} class:
\begin{equation}
    p(C_0|\mathbf{e}^v,\mathbf{e}^t_{y_0}) = \text{Sigmoid}(\text{Cosine\_Sim}(\mathbf{e}^v,\mathbf{e}^t_{y_0})).
\end{equation}
In Sec~\ref{sec:exp}, we will compare this adaptation and show its limitations, especially when tested on the proposed benchmark.

\subsection{Attribute Generation}
\label{m:AttGen}

As can be seen in Fig.~\ref{fig:method} (i), we utilize large language models to generate (text) candidates for functional and/or visual attributes. 
We prompt gpt-3.5 to list relevant attributes to identify the classes using the following template~\cite{ClsLLM, LOVM}:
\begin{align}
    Y_{C, Z } &= \parbox[t]{\dimexpr\linewidth-2cm}{
    I am using a language-vision model to identify \{C\}. List the \{Z\} attributes of \{C\}, which will be used for detection.}
    \label{eq:prompt}
\end{align}
With $C$ being the class name and $Z$ being the prompted attribute category (e.g., shape, size, texture, etc.).

We then generate a list of $N$ class-agnostic attributes using Alg.~\ref{algo:llm}, denoted by $A$, for attributes. To encode these attributes, we use the prompt template  ``\texttt{object which (is/has/etc) <Category> is <Attribute>}'', followed by the accepted prompt ensemble approaches with the model's text encoder~\cite{CLIP}.
This produces the attribute embedding matrix, $\mathbf{E}_{\text{att}} = [\mathbf{e}^t_{A_1}, \mathbf{e}^t_{A_2}, ... , \mathbf{e}^t_{A_N}]\in\mathbb{R}^{D\times N}$.
Now, a score between an object visual embedding $\mathbf{e}^v$ and an attribute embedding $\mathbf{e}^t_A$ is computed as follows:
\begin{equation}
    s_A = \text{Cosine\_Sim}(\mathbf{e}^v,\mathbf{e}^t_{A}).
    \label{eqn:s_A}
\end{equation}
We express the vector of attribute scores given an image embedding as $\mathbf{s} = [s_{A_1}, s_{A_2}, ... s_{A_N}]^\intercal \in \mathbb{R}^N$.

\subsection{Attribute Selection and Refinement}
\label{m:AttSelRef}
Not all the generated attributes are relevant to the problem at hand.
We use the training images of the known classes to select the relevant attributes.
We identify the relevant attributes by their predictive power for the known class classification.
To arrive at a class prediction from attributes, we introduce  the weight matrix, $\mathbf{W}\in\mathbb{R}^{K\times N}$.
Thus, the probability of a class $C$ can be computed given a attribute score vector $\mathbf{s}$:
\begin{equation}
p(C|\mathbf{s}) = \text{Sigmoid}(\mathbf{W}\mathbf{s}).
\end{equation}

To select the relevant attributes, we freeze the attribute embeddings, $\mathbf{E}_{\text{att}}$, and optimize $\mathbf{W}$ to classify the example images correctly (Fig.~\ref{fig:method}, ii).
We use the Binary-Cross Entropy (BCE) loss for optimization and $L_1$ regularization to encourage sparsity. 
We then select the top $\hat{N}$ attributes per object class and remove all the attributes not used in classifying any object class. 
To refine the attributes, we optimize $\mathbf{E}_{\text{att}}$ while freezing $\mathbf{W}$ (Fig.~\ref{fig:method}, iii) with the BCE loss. 

\begin{algorithm}[t]
\caption{Our LLM Attribute Generation.}
\label{algo:llm}
\textbf{Input:} $C_1, C_2, .. C_K$ \;
\textbf{Output:} $A$\_\text{list} \;
\textbf{Require:} \texttt{LLM}, $Y_{C,Z}$ Prompt Template (Eq.~\ref{eq:prompt})\;

$A$\_\text{list} $\leftarrow$ $\{\}$ \;
\For{ $C$ in  $C_1, C_2, .. C_K$ }{
   $A$\_\text{list}  $\leftarrow$ $A$\_\text{list} $ \cup\;$  $\texttt{LLM}(Y_C)$ \;
}
\KwRet $A$\_list \;
\end{algorithm}

\vspace{-0.1in}
\subsection{Unknown Class Inference}
\label{m:inference}
To identify an object as an \textit{unknown}, we want it to be \textit{in-distribution} with respect to the attributes, but \textit{out-of-distribution} with respect to the known classes (see Fig.~\ref{fig:method}, iv). Any out-of-distribution classification is suitable, and we select the simplest implementation -- max of softmax~\citep{MCM}:
\begin{equation}
p_{\text{OOD}} = 1 - \max_C\big(\text{SoftMax}_C(\mathbf{W}\mathbf{s})\big).  
\end{equation}
To determine if something is in distribution w.r.t. the attributes, we want to have high activation:
\begin{equation}
p_{\text{ID}} = \max_A\big(\text{Sigmoid}(\mathbf{s})\big),
\end{equation}
to produce the final unknown object prediction score, we simply compute
$p_\text{unk} = p_{\text{OOD}} \cdot p_{\text{ID}}$.

\section{Experiments}
\label{sec:exp}
We performed extensive experiments on the accepted open world object detection (OWD) benchmarks, comparing simple baselines (detailed below) to existing OWD methods (Sec.~\ref{sec:exp:owd}). We then compared these baselines to \ourMethod\ on the proposed RWD benchmark (Sec.~\ref{sec:exp:rowd}). 
We then study what attributes are selected and the relationship of the attributes to unknown objects (Sec.~\ref{sec:selected_attributes}).
To illustrate the effect of each of \ourMethod's components, we performed ablation experiments on the RWD benchmark (Sec.~\ref{sec:ablation}). 

%%%%%%%%%%%%%%%%%%%%%%%%% UPER SECTION %%%%%%%%%%%%%%%%%%%%%%%%%%%
%BASE-B/16
% Task 1
\newcommand{\BURTonesummart}{58.6}
\newcommand{\BCKTonesummart}{61.8}

%BASE-L/16
% Task 1
\newcommand{\LURTonesummart}{69.0}
\newcommand{\LCKTonesummart}{65.7}

%%%%%%%%%%%%%%%%%%%%%%%%% LOWER SECTION %%%%%%%%%%%%%%%%%%%%%%%%%%%

%%%%%%%%%%%%%%%%%%%%%%%%% UPER SECTION %%%%%%%%%%%%%%%%%%%%%%%%%%%
%BASE-B/16
% Task 1
\newcommand{\LBURTonesummart}{60.5}
\newcommand{\LBCKTonesummart}{59.5}

% Task 4
\newcommand{\LLURTonesummart}{69.3}
\newcommand{\LLCKTonesummart}{69.1}

\begin{table}[t]
\centering

\setlength{\tabcolsep}{6pt}
\adjustbox{width=\linewidth}{
\begin{tabular}{@{}l|cc|cccc|}
\toprule
 \textbf{Task IDs} ($\rightarrow$)
 & \multicolumn{2}{c|}{\textbf{Task 1}} 
 & \multicolumn{3}{c}{\textbf{Task 4}} 
\\ \midrule
 
& \cellcolor[HTML]{FFFFED}{U-Recall}&  \multicolumn{1}{c|}{\cellcolor[HTML]{EDF6FF}{mAP ($\uparrow$)}} & 
\multicolumn{3}{c}{\cellcolor[HTML]{EDF6FF}{mAP ($\uparrow$)}}   \\ 

 & \cellcolor[HTML]{FFFFED}($\uparrow$)
 &  \begin{tabular}[c]{@{}c}CK\end{tabular} 
 
 &  \begin{tabular}[c]{@{}c@{}}PK\end{tabular} 
 & \begin{tabular}[c]{@{}c@{}}CK\end{tabular} 
 & Both 
  \\ 
 
\midrule

OW-DETR~\citep{OWDETR} & \cellcolor[HTML]{FFFFED}7.5 & 59.2 & 31.4&  17.1 & 27.8  \\

CAT~\citep{CAT}  & \cellcolor[HTML]{FFFFED} 23.7 &   60.0 & 34.4 & 16.6 &  29.9 \\ 

PROB~\citep{PROB}  & \cellcolor[HTML]{FFFFED} 19.4 &   59.5 &  35.7 & 18.9 &  31.5  \\ 

Hyp-OW~\citep{doan2023hyp}  & \cellcolor[HTML]{FFFFED} 23.5  &   59.4  &  37.4 &  22.4 & 33.6 \\ 

MEPU-SS~\citep{fang2023unsupervised}  & \cellcolor[HTML]{FFFFED} 30.3 &   60.0  &34.3 & 18.9 & 30.4  \\

\midrule

MEPU-FS~\citep{fang2023unsupervised}  & \cellcolor[HTML]{FFFFED} 31.6 &  60.2  &  34.8 & 18.9 &  30.4  \\ 

MAVL~\citep{MAVL}  & \cellcolor[HTML]{FFFFED} 50.1 &  64.0  &   36.2 & 20.6 & 32.3 \\

\base-B/16   
& \cellcolor[HTML]{FFFFED}  \BURTonesummart &  \BCKTonesummart & 48.3 & 29.2 &  43.5 \\

\baseIN-B/16  & \cellcolor[HTML]{FFFFED}  63.3 & 61.5 &  48.7 &  29.8 &  44.0 \\

\baseLLM-B/16   
& \cellcolor[HTML]{FFFFED}  66.9 & 62.1 & 48.7 & 29.8 & 44.0 \\

\base-L/14 & \cellcolor[HTML]{FFFFED} 69.0 &  \textbf{65.7} &   57.2 &  50.5 &  55.5 \\

%%%%%%%%%%%%

\baseIN-L/14 &  \cellcolor[HTML]{FFFFED} 74.8 &  65.4 &   \textbf{57.5} &   \textbf{50.8} &   \textbf{55.8} \\

%%%%%%%%%%%%

\baseLLM-L/14 & \cellcolor[HTML]{FFFFED} \textbf{79.0} &  \textbf{65.7} &   \textbf{57.5} &  \textbf{50.8} &   \textbf{55.8} \\

\midrule

\baseGT-B/16$^\dagger$   
& \cellcolor[gray]{0.9}   71.5 &   \cellcolor[gray]{0.95}  62.1  &  \cellcolor[gray]{0.95} 48.7 & \cellcolor[gray]{0.95}  29.8  &  \cellcolor[gray]{0.95}  44.0

\\ 

\baseGT-L/14$^\dagger$
& \cellcolor[gray]{0.9} 85.7
 &   \cellcolor[gray]{0.95}  65.8
&   \cellcolor[gray]{0.95} 57.5

&  \cellcolor[gray]{0.95}  50.8 &  \cellcolor[gray]{0.95}  55.8
\\

\bottomrule

\end{tabular}
}
\vspace{-0.09in}
\caption{\textbf{Baseline results on M-OWODB.} Comparison of unknown class recall and mAP (U-R/U-mAP), and (previously/currently) known class mAP@$0.5$ of $0$-shot baselines to previous OWD methods. 
(top) Traditional OWD methods. (middle) Methods utilizing foundation models. (bottom) $^\dagger$GT baselines use the ground-truth class names to identify unknown objects, operating in the open-vocabulary paradigm, and serve as an upper bound for our baselines. 
For the full table (all tasks and S-OWODB), see Sup. Tab.~\ref{table:t1_owod}. 
}
\vspace{-0.1in}
\label{table:t1_owod_main}
\end{table}

\paragraph{Baselines.}
OVD methods still do not handle \emph{unknown} classes as they require class names to be given. We introduce several baselines that incorporate unknown object detection by augmenting OWL-ViT~\cite{owl-vit}:
\begin{itemize}
    \item \textbf{\base:}  Uses a generic prompt for unknown objects (``\texttt{object}''), in line with Maaz et al.~\cite{MAVL}.
    \item \textbf{\baseIN:} We use all of the ImageNet class names (after removing the known objects) as proposals for unknown objects.
    \item \textbf{\baseLLM:} Rather than just using a generic prompt, we use LLMs to predict possible unknown objects given the known classes. Then, we used the LLM-generated class names to detect unknown objects. 
    \item \textbf{\baseGT:} We use the ground-truth unknown object class names. This is the upper bound of the pure `open-vocabulary' approach, which assumes one has access to the ground truth of unknown class names. 
    \item \textbf{\baseFS:} All other baselines operate in the open-vocabulary/zero-shot detection setting. To compare to a baseline that receives the same supervision  (few-shot), we use the image exemplars to generate the vision-derived objects embeddings, which are per-class averaged to generate the class embeddings, as introduced in~\cite{owl-vit}). We then use a generic prompt to identify unknown objects (like ``\texttt{a photo of an object}''). As text-based embedding had lower cosine similarity than their image-derived counterparts, we selected unknown and known object predictions separately (50 from each per image). 
\end{itemize}

\paragraph{Implementation Details.}
\ourMethod\ is initialized using a frozen, CLIP pre-trained OWL-ViT (L/14 and B/16) model, which was detection fine-tuned on a federated dataset consisting of Objects 365 and Visual Genome~\citep{owl-vit}.
\ourMethod\ uses this frozen model and primarily deals with creating the best possible attribute embeddings while augmenting the inference pipeline to incorporate unknown object detection, as described in Sec.~\ref{m:inference}.

\paragraph{Evaluation Metrics.} For known classes, mean average precision (mAP) is used. 
To better understand the quality of continual learning, mAP is partitioned into previously and newly introduced object classes (Prev./Curr. Known). 
As common in OWD, we use unknown object recall (U-Recall), the ratio of detected to total labeled unknown objects~\citep{TowardsOWOD, OWDETR, PROB} on previous OWD benchmarks. Meanwhile, we report the unknown object mAP (U-mAP) on RWD.

\begin{table*}[t]
\centering
\vspace{-0.1in}
\setlength{\tabcolsep}{3pt}
\adjustbox{width=\textwidth}{
\Large
\begin{tabular}{@{}ll|cc|cc|cc|cc|cc|cc|cc|cc|cc|cc||cc|cc@{}}
\toprule
  \multicolumn{2}{c|}{\textbf{Domain}} & \multicolumn{4}{c|}{\textbf{Aquatic}} & \multicolumn{4}{c|}{\textbf{Aerial}} & \multicolumn{4}{c|}{\textbf{Game}}& \multicolumn{4}{c|}{\textbf{Medical}} & \multicolumn{4}{c||}{\textbf{Surgery}}&  \multicolumn{4}{c}{ \textbf{Overall}} \\ 
 
 \midrule
 \multicolumn{2}{c|}{ \textbf{Task IDs} ($\rightarrow$)} & \multicolumn{2}{c|}{\textbf{Task 1}} & \multicolumn{2}{c|}{\textbf{Task 2}} & \multicolumn{2}{c|}{\textbf{Task 1}} & \multicolumn{2}{c|}{\textbf{Task 2}} 
& \multicolumn{2}{c|}{\textbf{Task 1}} & \multicolumn{2}{c|}{\textbf{Task 2}} 
& \multicolumn{2}{c|}{\textbf{Task 1}} & \multicolumn{2}{c|}{\textbf{Task 2}} 
& \multicolumn{2}{c|}{\textbf{Task 1}} 
& \multicolumn{2}{c||}{\textbf{Task 2}}
& \multicolumn{2}{c|}{\textbf{Task 1}} & \multicolumn{2}{c}{\textbf{Task 2}}\\

  &&
  \cellcolor[HTML]{FFFFED} U & \cellcolor[HTML]{EDF6FF} K & \cellcolor[HTML]{EDF6FF} PK & \cellcolor[HTML]{EDF6FF} CK &
  \cellcolor[HTML]{FFFFED} U & \cellcolor[HTML]{EDF6FF} K & \cellcolor[HTML]{EDF6FF} PK & \cellcolor[HTML]{EDF6FF} CK &
  \cellcolor[HTML]{FFFFED} U & \cellcolor[HTML]{EDF6FF} K & \cellcolor[HTML]{EDF6FF} PK & \cellcolor[HTML]{EDF6FF} CK &
  \cellcolor[HTML]{FFFFED} U & \cellcolor[HTML]{EDF6FF} K & \cellcolor[HTML]{EDF6FF} PK & \cellcolor[HTML]{EDF6FF} CK &
  \cellcolor[HTML]{FFFFED} U & \cellcolor[HTML]{EDF6FF} K & \cellcolor[HTML]{EDF6FF} PK & \cellcolor[HTML]{EDF6FF} CK &
  \cellcolor[HTML]{FFFFED} U & \cellcolor[HTML]{EDF6FF} K & \cellcolor[HTML]{EDF6FF} PK & \cellcolor[HTML]{EDF6FF} CK 
\\

\midrule

%%%%%%
 & \baseGT-B/16$^\dagger$
 & \cellcolor[gray]{0.9} 29.8 
 & \cellcolor[gray]{0.95} 45.0 
 & \cellcolor[gray]{0.95}45.0 
 & \cellcolor[gray]{0.95}36.7 
 &  \cellcolor[gray]{0.9}  1.3 
 & \cellcolor[gray]{0.95}5.7 
 & \cellcolor[gray]{0.95}5.7 
 & \cellcolor[gray]{0.95}1.4 
 &  \cellcolor[gray]{0.9}  15.0 
 & \cellcolor[gray]{0.95}0.4
 & \cellcolor[gray]{0.95}0.4 
 & \cellcolor[gray]{0.95}0.1 
 &  \cellcolor[gray]{0.9}  0.5 
 & \cellcolor[gray]{0.95}0.0 
 & \cellcolor[gray]{0.95}0.0 
 & \cellcolor[gray]{0.95}0.1 
 &  \cellcolor[gray]{0.9}  5.6
 & \cellcolor[gray]{0.95}1.5 
 & \cellcolor[gray]{0.95}1.4 
 & \cellcolor[gray]{0.95}0.3 
 &  \cellcolor[gray]{0.9}  10.4
 & \cellcolor[gray]{0.95}10.5 
 & \cellcolor[gray]{0.95}10.5
 & \cellcolor[gray]{0.95}7.7 \\

  & \baseGT-L/14$^\dagger$
& \cellcolor[gray]{0.9}  34.8 
& \cellcolor[gray]{0.95} 36.0 
& \cellcolor[gray]{0.95} 36.0 
& \cellcolor[gray]{0.95} 42.3 
&  \cellcolor[gray]{0.9}  1.0 
& \cellcolor[gray]{0.95} 7.9 
& \cellcolor[gray]{0.95} 7.2 
& \cellcolor[gray]{0.95} 0.8 
&  \cellcolor[gray]{0.9}  12.4 
& \cellcolor[gray]{0.95} 0.9 
& \cellcolor[gray]{0.95} 0.8 
& \cellcolor[gray]{0.95} 0.3 
&  \cellcolor[gray]{0.9}  2.4 
& \cellcolor[gray]{0.95} 0.2 
& \cellcolor[gray]{0.95} 0.2
& \cellcolor[gray]{0.95} 0.3 
&  \cellcolor[gray]{0.9}  2.4 
& \cellcolor[gray]{0.95} 0.2 
& \cellcolor[gray]{0.95} 2.6 
& \cellcolor[gray]{0.95} 1.3 
&  \cellcolor[gray]{0.9}  10.6 
& \cellcolor[gray]{0.95} 9.0 
& \cellcolor[gray]{0.95} 9.4
& \cellcolor[gray]{0.95} 9.0 \\

\midrule

 & \base-B/16
& \cellcolor[HTML]{FFFFED} 6.2 & 45.0 & 45.0 & 36.7 & \cellcolor[HTML]{FFFFED} 0.9 & 5.7 & 5.7 & 1.4 &  \cellcolor[HTML]{FFFFED} 15.7 & 0.4 & 0.4 & 0.1 &  \cellcolor[HTML]{FFFFED} 0.2 & 0.0 & 0.0 & 0.1 &  \cellcolor[HTML]{FFFFED} 1.4 & 1.5 & 1.4 & 0.3 &  \cellcolor[HTML]{FFFFED} 4.9 & 10.5 & 10.5 & 7.7 \\

 & \baseIN-B/16
& \cellcolor[HTML]{FFFFED} 26.5 & 45.1 & 45.1 & 36.7 & \cellcolor[HTML]{FFFFED} 1.9 & 5.7 & 5.7 & 1.4 & \cellcolor[HTML]{FFFFED} 2.4 & 0.3 & 0.3 & 0.0 &  \cellcolor[HTML]{FFFFED} 0.6 & 0.0 & 0.0 & 0.1 &  \cellcolor[HTML]{FFFFED} 1.7 & 1.4 & 1.0 & 0.3 &  \cellcolor[HTML]{FFFFED} 6.6 & 10.5 & 10.4 & 7.7 \\

 & \baseLLM-B/16 
& \cellcolor[HTML]{FFFFED} 24.7 & 45.1 & 45.1 & 36.5 &  \cellcolor[HTML]{FFFFED} 1.4 & 5.7 & 5.7 & 1.4 &  \cellcolor[HTML]{FFFFED} 15.1 & 0.4 & 0.4 & 0.1 &  \cellcolor[HTML]{FFFFED} 0.6 & 0.0 & 0.0 & 0.1 & \cellcolor[HTML]{FFFFED}  8.9 & 1.5 & 1.3 & 0.3 & \cellcolor[HTML]{FFFFED} 10.2 & 10.5 & 10.5 & 7.7 \\

 & \base-L/14 
& \cellcolor[HTML]{FFFFED} 0.7 & 35.9 & 36.0 & 42.3 &  \cellcolor[HTML]{FFFFED} 9.1 & 8.2 & 7.2 & 0.8 & \cellcolor[HTML]{FFFFED} 6.8 & 0.9 & 0.8 & 0.3 & \cellcolor[HTML]{FFFFED} 0.0 & 0.2 & 0.2 & 0.3 & \cellcolor[HTML]{FFFFED} 3.6 & 2.9 & 2.6 & 1.3 &  \cellcolor[HTML]{FFFFED} 4.1 & 9.6 & 9.4 & 9.0 \\

 & \baseIN-L/14  & \cellcolor[HTML]{FFFFED} 19.6 & 35.8 & 35.8 & 41.8 &  \cellcolor[HTML]{FFFFED} 2.3 & 7.2 & 6.9 & 0.9 &  \cellcolor[HTML]{FFFFED} 15.8 & 0.9 & 0.8 & 0.3 &  \cellcolor[HTML]{FFFFED} 0.9 & 0.1 & 0.1 & 0.2 & \cellcolor[HTML]{FFFFED} 3.1 & 2.1 & 1.9 & 1.1 &  \cellcolor[HTML]{FFFFED} 8.3 & 9.2 & 9.1 & 8.8 \\

 %%%%%%

 & \baseLLM-L/14 
& \cellcolor[HTML]{FFFFED} 24.7 & 35.8 & 35.8 & 42.2 & \cellcolor[HTML]{FFFFED} 0.6 & 7.6 & 7.2 & 0.8 &  \cellcolor[HTML]{FFFFED} 12.5 & 0.9 & 0.8 & 0.2 & \cellcolor[HTML]{FFFFED} 1.6 & 0.1 & 0.1 & 0.2 & \cellcolor[HTML]{FFFFED}  12.6 & 2.6 & 2.5 & 1.3 & \cellcolor[HTML]{FFFFED} 10.4 & 9.4 & 9.3 & 9.0 \\

  %%%%%%

 & \baseFS-B/16 
 & \cellcolor[HTML]{FFFFED} 7.1  & 41.1 & 41.1 & 31.9 
 & \cellcolor[HTML]{FFFFED} 1.2  & 10.4 & 10.1 & 4.0 
 & \cellcolor[HTML]{FFFFED} 16.0 & 4.6  & 4.8  & 3.9  
 & \cellcolor[HTML]{FFFFED} 0.6  & 6.1  & 6.1  & 3.3  
 & \cellcolor[HTML]{FFFFED} 1.3  & 11.9 & 11.3 & 10.9 
 & \cellcolor[HTML]{FFFFED} 5.2  & 14.8 & 14.7 & 10.8 \\

 & \baseFS-L/14 
 & \cellcolor[HTML]{FFFFED} 2.4 & 43.6 & 42.9 & 42.8 
 & \cellcolor[HTML]{FFFFED} 9.7 & 23.7  & 21.9& 13.0 
 & \cellcolor[HTML]{FFFFED} 8.2 & 10.4 & 10.2 & 13.4 
 & \cellcolor[HTML]{FFFFED} 1.1 & 23.2 &  21.7 & 24.2 
 & \cellcolor[HTML]{FFFFED} 3.6 & 26.0 & 25.0 & 7.4  
 & \cellcolor[HTML]{FFFFED} 5.0 & 25.4 & 24.3 & 20.2 \\

 %%%%%%%%%
 & \ourMethod-B/16 
 & \cellcolor[HTML]{FFFFED} 3.5 & 43.8 & 44.1 & 40.8
 & \cellcolor[HTML]{FFFFED} 0.9 & 12.0 & 12.6 & 5.4 
 & \cellcolor[HTML]{FFFFED} 13.3 & 3.8 & 4.4 & 4.1  
 & \cellcolor[HTML]{FFFFED} 2.1 & 6.4 & 5.5 & 11.5
 & \cellcolor[HTML]{FFFFED} 6.1 & 12.7 & 12.9 &  11.0
 & \cellcolor[HTML]{FFFFED} 5.2 & 15.7 & 15.9 & 14.6 \\

 & \ourMethod-L/14 
 & \cellcolor[HTML]{FFFFED}  18.2 & 50.1 & 48.1 & 47.1
 & \cellcolor[HTML]{FFFFED} 6.0 & 25.3 & 23.7 & 16.0 
 & \cellcolor[HTML]{FFFFED} 30.4 & 10.7 & 9.9 & 11.2
 & \cellcolor[HTML]{FFFFED} 9.4 & 21.8 & 19.9 &  34.6
 & \cellcolor[HTML]{FFFFED} 12.0 &  29.0 &  28.9 & 8.5
 & \cellcolor[HTML]{FFFFED} \textbf{15.2} & \textbf{27.4} & \textbf{26.1} & \textbf{23.5} \\

\bottomrule
\end{tabular}
}
\vspace{-0.09in}
\caption{\label{table:t2_rwod_new}
\textbf{RWD benchmark.} 
Each domain is broken up into two tasks, where in task 1, you are given some classes, and then known and unknown mAP are evaluated. In Task 2, the rest of the classes are revealed, and we evaluate Previously and Currently known mAP.
\ourMethod\ and \baseFS\ are evaluated in 100-shot regime. To see the effect of the number of shots on performance, please see Sup. Sec.~\ref{sup:sec:fs_exp}.
$^\dagger$GT baselines use the ground-truth class names to identify unknown objects, effectively operating in the open-vocabulary paradigm, and serve as an upper bound to the text-conditioned (zero-shot) baselines.
}
\end{table*}

\begin{figure*}
\vspace{-0.1in}
  \centering
  \begin{subfigure}{0.19\textwidth}
  \centering
  Aquatic
  \end{subfigure}
  \begin{subfigure}{0.17\textwidth}
  \centering
  Arial
  \end{subfigure}
  \begin{subfigure}{0.22\textwidth}
  \centering
  Game
  \end{subfigure}
  \begin{subfigure}{0.16\textwidth}
  \centering
  Medical
  \end{subfigure}
  \begin{subfigure}{0.22\textwidth}
  \centering
  Surgery
  \end{subfigure}
  \centering
  \includegraphics[width=1\textwidth]{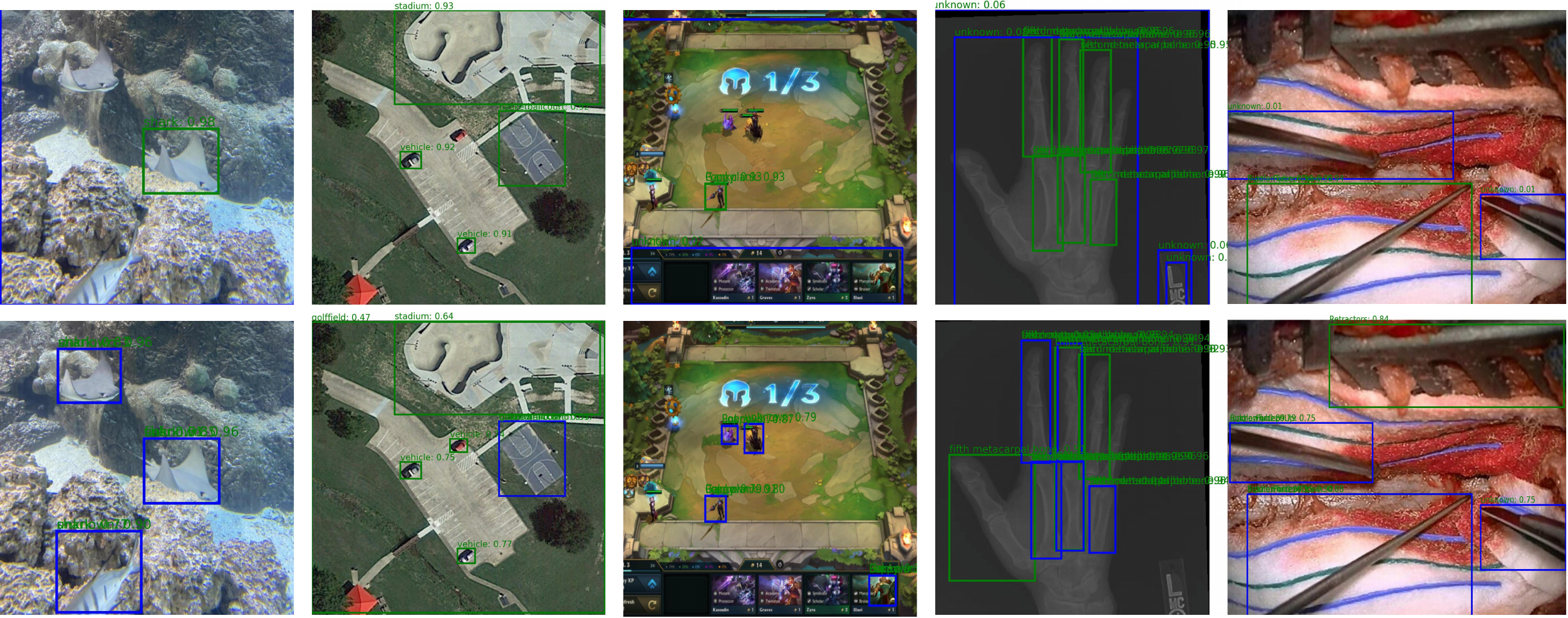}
    \vspace{-0.2in}
  \caption{\textbf{Qualitative Results.} Qualitative results (top) \baseFS\ (bottom) \ourMethod\ on RWD. {\color{blue}blue} - unknown, {\color{green}green} - known.
  \ourMethod\ shows superior performance on RWD, appearing to have less known-class confusion and better unknown object detection capability. For example, in Aquatic, \baseFS\ seems to confuse unknown objects (stingrays) as known objects (sharks), but \ourMethod\ appears more robust. 
  }
  \vspace{-0.14in}
\label{fig:qualitative}
\end{figure*}

\paragraph{Datasets.} 
Please see Sec.~\ref{sec:rwod} for a description of the RWD dataset.
Simple baselines were evaluated on the  accepted OWD benchmarks: ``superclass-mixed'' and ``superclass-separated'' OWD benchmarks (M-OWODB~\citep{TowardsOWOD} and S-OWODB~\citep{OWDETR}, respectively). 
In M-OWODB, images from MS-COCO~\citep{MSCOCO}, PASCAL VOC2007, and PASCAL VOC2012~\citep{VOC2007} are grouped into four sets of non-overlapping Tasks; $\{T_1,\cdots,T_4\}$ {s.t.} classes in a task $T_t$ are not introduced until $t$ is reached. 
In each task, $T_t$, an additional 20 classes are introduced - and in training for task $t$, only these classes are labeled, while in the test set, all the classes encountered 
need to be detected.
Only the MS-COCO dataset was used to construct S-OWODB, and a clear separation of super-categories (e.g., animals/vehicles) was performed. 
For more, please refer to~\cite{TowardsOWOD} and~\citet{OWDETR}.

\subsection{Foundation Model Baselines on Existing Open World Object Detection Benchmarks}
\label{sec:exp:owd}
Tab.~\ref{table:t1_owod_main} shows the performance of simple baselines utilizing foundation models on M-OWODB. 
Methods utilizing foundation models (including \base\ and MAVL~\cite{MAVL}) have high U-Recall ($>50$) across both benchmarks (See Tab.~\ref{table:t1_owod}). 
\base-L/14 outperforms methods not utilizing foundation models by $\sim2\times$ in terms of U-Recall and final known mAP. \baseGT\ serves as an upper bound to performance and operates entirely in the `open-vocabulary' definitions, where the class names of unknown objects are given to the model. Meanwhile, \baseLLM\ utilizes a large language model to generate candidate unknown object proposals, resulting in improvements to performance. 
The known mAP of all baselines is quite similar and is to be expected as all these models differ only in unknown object detection. 
As the performance variance between the LLM and GT models is marginal ($6.7$ U-Recall and $0.1$ K-mAP), more challenging benchmarks are needed to evaluate OWD methods that utilize foundation models. 

Yet, we emphasize that this does not invalidate prior OWD methods as they are trained from scratch on the OWD benchmarks. 
Our baselines, incorporating foundation models, inevitably will have seen objects (including the `unknown' objects) during the pre-training phase. 
So, the results are not directly comparable. 
We merely show that foundation model baselines can perform extraordinarily well in the existing COCO-based benchmarks, necessitating more challenging benchmarks for evaluation. 
Considering foundation models' impressive performance, we believe their utilization in OWD is well-motivated
For the full OWD results with both M- and S-OWODB, please see Sup. Tab.~\ref{table:t1_owod}, Tab.~\ref{tab:wi_ose} and Sec.~\ref{sup:sec:owod}.

\begin{figure}[t]
  \centering
  \includegraphics[width=0.95\linewidth]{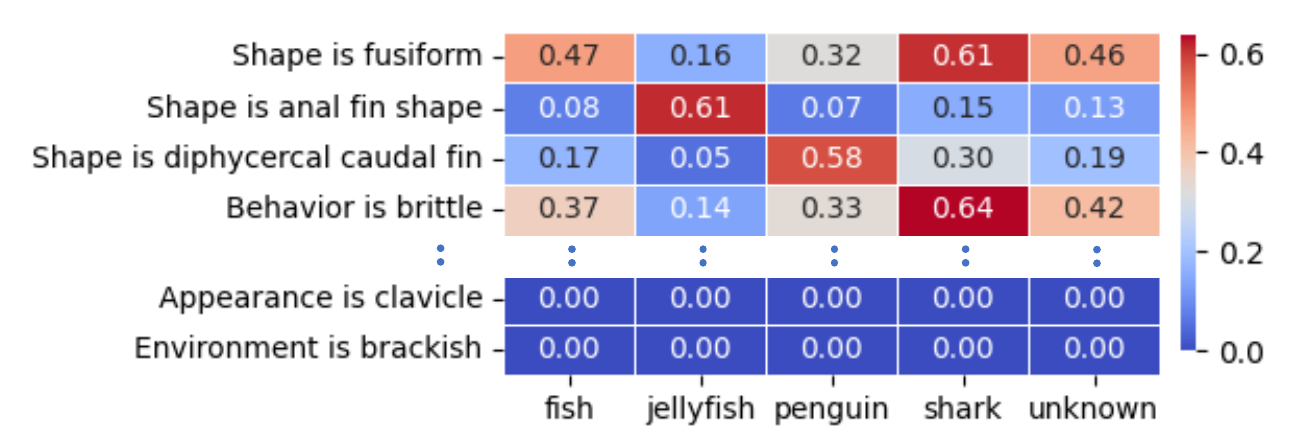}
  \vspace{-0.13in}
  \caption{
  \textbf{Attributes on Aquatic.}
  We selected the top attribute per known class 
  and two random attributes that were not selected. 
  }\vspace{-0.1in}
\label{fig:att}
\end{figure}

\subsection{Real-World Object Detection}
\label{sec:exp:rowd}
Next, we evaluate the proposed baselines alongside \ourMethod\ on the proposed benchmark, RWD. 

\paragraph{Quantitative Results.}
Tab.~\ref{table:t2_rwod_new} shows the performance of the different baselines and \ourMethod\ on the proposed RWD benchmark. \base\ had a poor zero-shot performance on the benchmark, indicating that RWD is out-of-distribution for this foundation model. \baseGT\ completely adopts the open-vocabulary object detection paradigm and even assumes access to the ground-truth unknown class names (which, realistically, one does not have access to), therefore serving as an upper bound to these approaches. Nonetheless, due to the domain/content shifts in the benchmark, this approach fails with near-zero mAP across most of the benchmark. 
The strongest baseline, \baseFS\, utilizes the few-shot adaptation approach introduced by Minderer et al.,~\cite{owl-vit}. However, as there is no supervision for the `unknown' object class, we had to utilize text-based detection for this class. As a result, we had to select unknown and known object proposals separately, as the scaling of text- and image-based embeddings was not comparable. 
Compared to \baseFS, \ourMethod\ outperformed with an $\sim3\times$ increase in U-mAP and a smaller (but still significant) improvement of $\sim20\%$ in known mAP.
Our results demonstrate the validity of utilizing attributes to detect known and unknown objects.

\paragraph{Qualitative Results.} Fig.~\ref{fig:qualitative} showcases the qualitative showcases of \ourMethod\ and \baseFS\ on RWD, with \ourMethod\ exhibiting superior performance. In the Surgery application domain, while \baseFS\ misses the Bipolar Forceps, \ourMethod\ correctly identifies it. 
Furthermore, \ourMethod's unknown object predictions seem reasonable, for example, identifying characters in the Game or bones in the Medical datasets.

\subsection{Selected Attributes Study\label{sec:selected_attributes}}
Next, we were interested in studying the relationship between the attribute scores and the known and unknown class predictions. Specifically, in the Aquatic dataset, we selected the top-scoring attribute per class (over the entire dataset), and two attributes that were not selected during Attribute Selection, and show the results in Fig.~\ref{fig:att}. We can see that some attributes make a lot of sense; for example, for `Fish': shape is fusiform (tapering at both ends), which generally describes most fish. Interestingly, we can see that the unknown objects exhibit relatively high scores for these attributes, hinting that, as expected, the unknown objects are in-distribution w.r.t. these attributes. We also selected (at random) two attributes that were not selected. At first glance, it makes some sense why they were not selected -- in the Aquatic dataset, classifying different animals is not affected by whether or not the environment is brackish, for example.

\subsection{Ablation Study\label{sec:ablation}}
\begin{table}[t]
\centering
\setlength{\tabcolsep}{3pt}
\adjustbox{width=0.6\linewidth}{
\begin{tabular}{@{}l|cc|cc@{}}
\toprule
 \textbf{Task IDs} ($\rightarrow$) & \multicolumn{2}{c|}{\textbf{Task 1}} & \multicolumn{2}{c}{\textbf{Task 2}} \\
   & \cellcolor[HTML]{FFFFED} U & \cellcolor[HTML]{EDF6FF} K & \cellcolor[HTML]{EDF6FF} PK & \cellcolor[HTML]{EDF6FF} CK\\ 
    \midrule

\ \ \  \ \hfill \ourMethod & \cellcolor[HTML]{FFFFED} 15.2 & 27.4 & 26.1 & 23.5 \\
\ \ \ $-$ Refinement       &\cellcolor[HTML]{FFFFED}  15.1 & 24.9 & 24.2 & 20.7 \\
\ \ \ $-$ Selection        & \cellcolor[HTML]{FFFFED}  15.7 & 9.6 & 3.5 & 7.0 \\
\bottomrule
\end{tabular}
}
\vspace{-0.07in}
\caption{\textbf{\ourMethod\ Ablation on RWD.} 
We show the effect of progressively integrating our contributions on $100$-shot \ourMethod-L/14. We report U, K, PK, and CK-mAP on RWD.}
\label{table:ablation}
\vspace{-0.1in}
\end{table}

 Tab.~\ref{table:ablation} displays the effects of sequentially removing our contributions. Removing attribute refinement leads to an average decrease in known object mAP of $2.4$. 
 Attribute selection significantly improves model performance, showing the importance of image-based attribute selection; without it, the average known mAP drops by $19.0$.
 On the other hand, unknown mAP seems largely unaffected by these changes, with a $0.1$ drop and $0.5$ improvement when removing refinement and selection, respectively. However, when comparing unknown mAP to \baseFS, it is clear that utilizing attributes benefits unknown object detection.
 Overall, our ablations showcase the necessity of each part of the proposed pipeline.

\noindent
For a discussion about limitations, please see Sup. Sec.~\ref{sec:limitations}. 

\section{Conclusions}

In this paper, we introduce foundation models to the OWD task. 
Initial baselines showed strong performance on the existing benchmarks, indicating the need for more challenging benchmarks when evaluating foundation models. Therefore, we curated a challenging benchmark, RWD, consisting of five datasets from different domains and high-impact applications that challenge these methods. We then introduced \ourMethod, which leveraged known object attributes to identify potential unknown objects, resulting in an impressive $\sim 3 \times$ gain in U-mAP.
At a broader level, our study presents the first work in integrating foundation VLMs into the open world object detection task. Although our findings demonstrate improved results, they also show that this task is far from being completely resolved.  We anticipate many avenues for future research. How do we fully harness the potential of foundation models for object detection while training it on a much smaller detection dataset?  Additionally, how do we adapt these models in the face of distribution shifts -- both in terms of new concepts and unseen styles?  We envision that this new object detection paradigm will accelerate the adoption into real-world domains and applications.

\textbf{Acknowledgments.} This project was funded by HAI-Google. We thank the Knight-Hennessy Scholars Foundation for generously funding OZ.

\newpage

\appendix
\clearpage
\setcounter{page}{1}
\maketitlesupplementary

\section{Real-World Object Detection Benchmark Details\label{sup:sec:benchmark_details}}
\begin{figure*}[t]
  \centering
  \includegraphics[width=\textwidth]{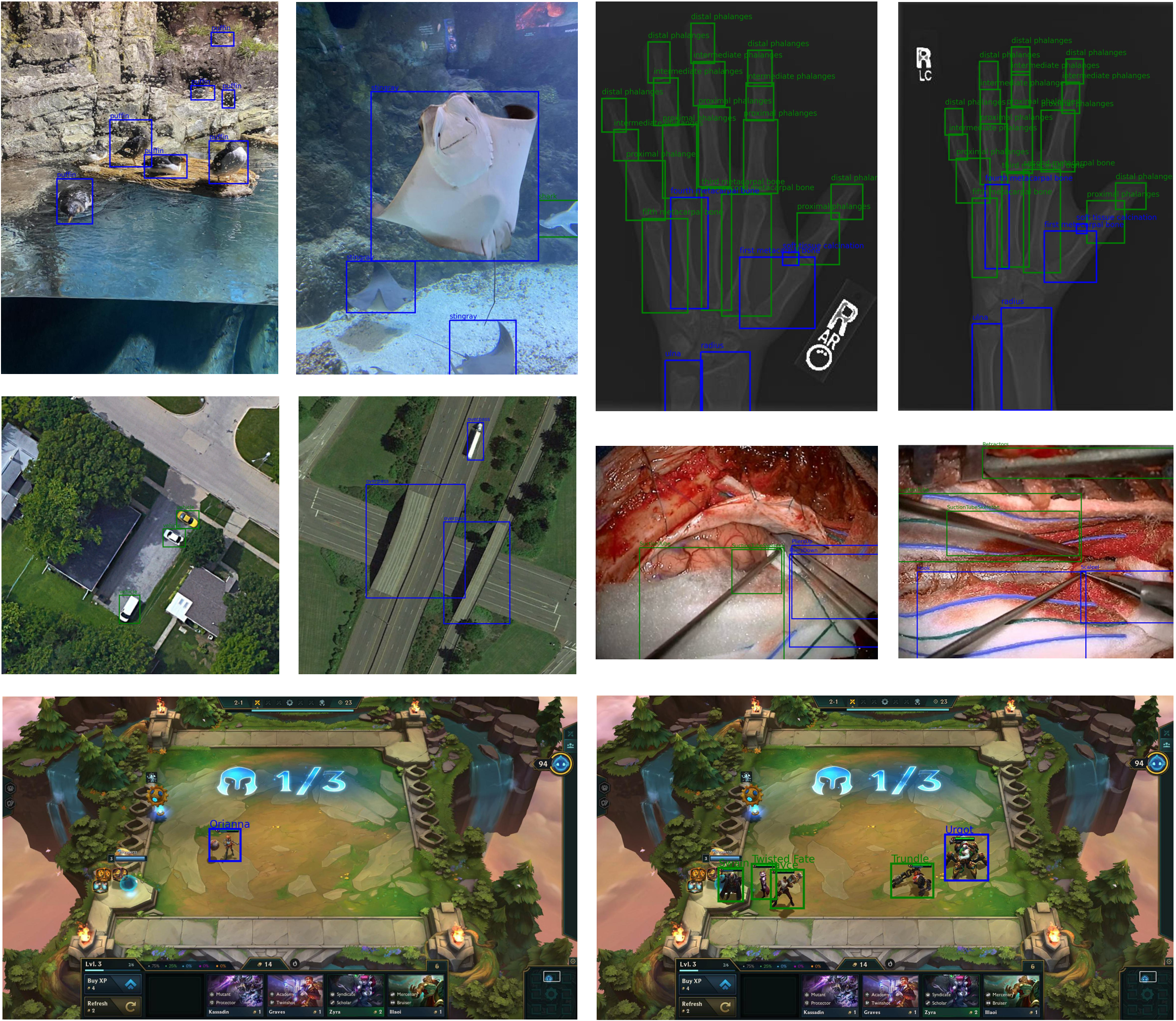}

  \caption{\textbf{RWD Dataset Visualization.} 
  Colors are used to denote {\color{green}\textit{Task 1 known (green)}} and {\color{blue} \textit{Task 1 unknown (blue)}}.
  We show the ground-truth objects for all the domains in RWD, specifically
  (top-left) Aquatic, (mid-left) Aerial, (bottom) Game, (top-right) Medical, and (mid-right) Surgical datasets. 
  The great diversity of RWD is apparent, showcasing the strength of the proposed benchmark.
  }
\label{fig:sup_benchmark}
\end{figure*}

\begin{table}
\centering
\adjustbox{width=\linewidth}{
\begin{tabular}{ c|ccc|c }
\toprule
Name &  \begin{tabular}{@{}c@{}}\# Train \\ Images \end{tabular} & \begin{tabular}{@{}c@{}}\# Test \\ Images \end{tabular} & \# Classes & \begin{tabular}{@{}c@{}}\% in \\ Tokenizer \end{tabular} \\
\toprule
Aquatic  &   318  &  319  &  7   &  100    \\
Aerial   &  5000  &  5000 &  20  &  55.0  \\
Game     &   788  &  787  &  59  &  35.6   \\
Medical  &   93   &   89  &  12  &  15.4    \\
Surgery  &   912  &  917  &  13  &  7.69  \\
\bottomrule
\end{tabular}}
\caption{\textbf{Datasets Details.} Number of classes and images, as well as what \% of the class names are in the tokenizer's vocabulary.}

\label{table:dataset_details_ood}
\end{table}
The Real-World Object Detection (RWD) benchmark assesses object detectors' ability to handle known and unknown objects in diverse domains.
Here, we detail the five object detection datasets we included, each representing a different domain. 
The Aquatic, Game, and Medical datasets were all taken from RoboFlow100's~\cite{roboflow100}, aquarium, team fight tactics, and xray-rheumatology datasets, respectively. The Aquatic dataset consists of underwater images of different sea creatures. The Game dataset consists of game snapshots containing different avatars. The Medical dataset consists of hand X-ray images where different bones are to be detected.  
The Aerial dataset was taken from DIOR~\citep{DIOR2020} and consisted of aerial images of stadiums/storage containers/ships/etc. The Surgery dataset was taken from the NeuroSurgicalTools dataset~\citep{NeuroSurgicalToolsDataset}, captured by neurosurgical microscopes, and contains various surgical tools. Tab.~\ref{table:dataset_details_ood} details the number of classes, train, and test images per dataset.
Fig.~\ref{fig:sup_benchmark} demonstrates example images from RWD across all domains and their ground truth bounding box annotations. According to the RWD benchmark class name breakdown for Task 1 known (green) and unknown (blue) classes, we use color coding to differentiate between classes.
These images provide a glimpse into the diverse visual content and object instances present in each sub-domain, offering a visual context for understanding the challenges and characteristics of RWD.

For our benchmark, we divided each dataset into two subsets: the 50\% most and least common classes. 
This split is well motivated by long-tailed recognition, which splits the classes by their scarcity~\cite{LongTailOW, LongTailOW_2}.
Please see Tab.~\ref{tab:dataset_class_names} for each dataset's class breakdown of the known/unknown objects.  
We evaluate object detection methods in two stages: Task 1 and Task 2. In Task 1, only the 50\% most common classes are considered known, while the 50\% least common classes remain unknown to the models. 
Detectors are expected to detect both known and unknown objects during this stage. 
This evaluation stage tests the ability of object detectors to detect novel objects and generalization to unknown classes. 
In Task 2, the remaining 50\% least common classes are revealed, and object detectors are evaluated based on their overall performance on the complete set of classes (Previously and Currently Known separately). Please see Tab.~\ref{tab:dataset_class_names}, which provides comprehensive information about the number of classes and the division between known and unknown classes in Task 1.

\begin{table*}[ht]
  \centering
  \begin{tabular}{c|p{5.7cm}|p{5.5cm}}
    \toprule
    Dataset & Task 1 Known / Task 2 Prev. Known & Task 1 Unknown / Task 2 Curr. Known\\
    \midrule
    Aquatic & Fish, Jellyfish, Penguin, Shark. & Puffin, Stingray, Starfish.\\

\vspace{-0.1in}&\vspace{-0.1in}&\vspace{-0.1in}\\
       
    Aerial & Airplane, Airport, Basketball Court, Dam, Golf Field, Ground Track Field, Ship, Stadium, Storage Tank, Vehicle. & Expressway-Service-Area, Expressway-Toll-Station, Baseball Field, Bridge, Chimney, Harbor, Overpass, Tennis Court, Train Station, Windmill.\\

\vspace{-0.1in}&\vspace{-0.1in}&\vspace{-0.1in}\\
    
    Game &  Singed, Ezreal, Janna, Vi, Tristana, Trundle, Gankplank, Blitzcrank, Heimerdinger, Swain, Warwick, Vex, Ziggs, Zilean, Jayce, Zac, Poppy, Darius, Twitch, Cho-Gath, Katarina, Braum, Twisted Fate, Kassadin, Lulu, Malzahar, Veigar, Caitlyn, Camille, Illaoi. & Talon, Lux, Seraphine, Jhin, Taric, Leona, Viktor, Lissandra, Yuumi, Akali, Ekko, Samira, Kai-Sa, Dr- Mundo, Fiora, Orianna, Jinx, Yone, Quinn, Miss Fortune, Sion, Kog-Maw, Garen, Graves, Urgot, Galio, Shaco, Zyra, Tahm Kench.\\

\vspace{-0.1in}&\vspace{-0.1in}&\vspace{-0.1in}\\

    Medical & Distal Phalanges, Proximal Phalanges, Intermediate Phalanges, Second Metacarpal Bone, Third Metacarpal Bone, Fifth Metacarpal Bone. & Soft Tissue Calcination, Ulna, Fourth Metacarpal Bone, Artifact, First Metacarpal Bone, Radius.\\

\vspace{-0.1in}&\vspace{-0.1in}&\vspace{-0.1in}\\
    
    Surgery & Suction Tube, Suction Tube Skeleton, Bipolar Forceps Up, Bipolar Forceps Down, Bipolar Forceps Up Skeleton, Retractors. & Curette, Hook, Scalpel, Bipolar Forceps Down,  Scissors, Pliers Down, Pliers Up.\\
    \bottomrule
  \end{tabular}
    \caption{\textbf{Real-World Object Detection Benchmark Class Name Breakdown.} The classes selected as Task 1 known/unknown and Task 2 previously/currently known objects are listed below. }
\label{tab:dataset_class_names}
\end{table*}

\section{Additional Implementation Details}
\label{sup:additional_implementation_details}
\ourMethod's base is the OWL-ViT model released by the Transformers library\footnote{\url{https://huggingface.co/docs/transformers/v4.28.1/en/model_doc/owlvit}}, which we augmented for our purposes. For few-shot detection, we input an image and the corresponding ground truth bounding box to the model and produced the predicted bounding box and class embeddings. We filtered the class embeddings by their corresponding bounding box such that all the class embeddings have an intersection over union of at least 0.8 with the ground truth object. We then selected the class embedding farthest from the mean of all the filtered class embeddings to produce the final image-condition class embedding.

\paragraph{Using Large Language Models to Discover Novel Classes.} The \baseLLM\ model augments a standard open-vocabulary detector by utilizing large language models to generate unknown object proposals.
Large language models (LLMs) have not been used to identify novel objects. Still, their ability to learn complex patterns and generate new sequences of words makes them well-suited for this task when used with multi-modal models. Specifically, we prompt OpenAI's DaVinci, with a temperature of 0.5:
\begin{figure}[H]
  \centering
  \includegraphics[width=\linewidth]{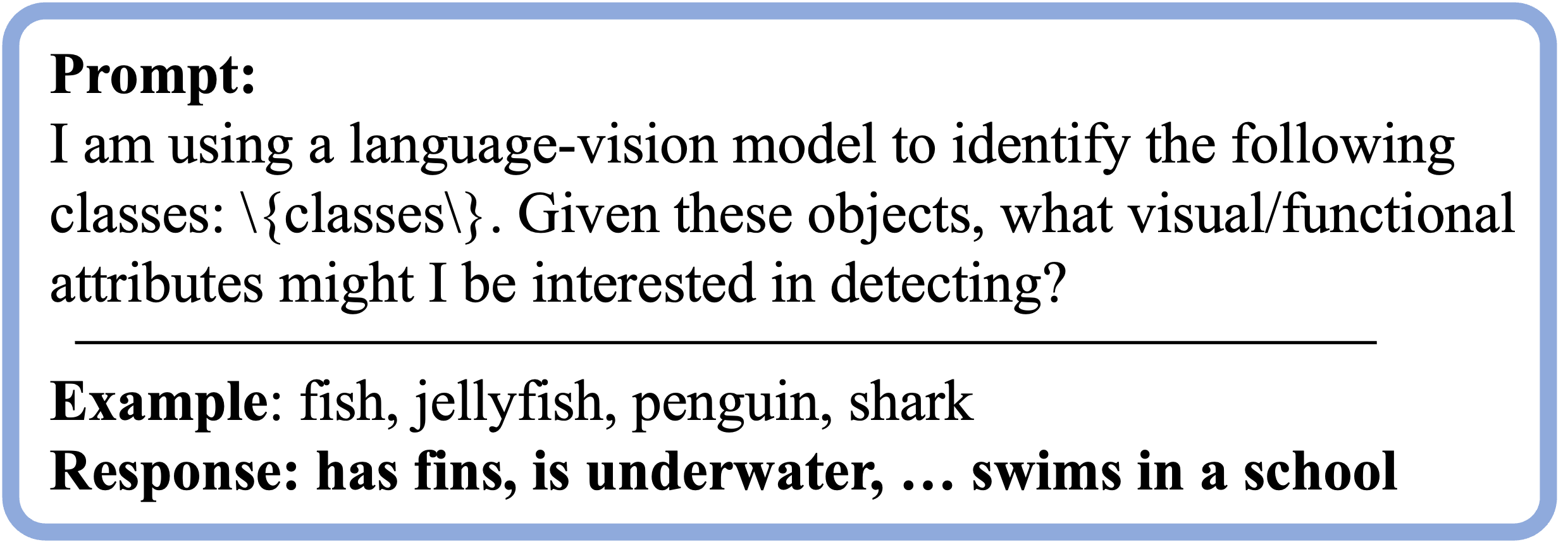}
\end{figure}
The LLM then produces a list of possible classes of interest (see Sup. Tab.~\ref{tab:llm_proposal}). 
These proposed classes are then used to augment the open-vocabulary detector to detect these classes as possible `unknowns'. Using smaller/chat LLMs resulted in poor unknown object proposals. 

LLM sometimes propose known object classes, which we filter out to mitigate known-unknown confusion.
Examples of novel object proposals can be seen in Tab.~\ref{tab:llm_proposal}. The table shows that the LLM adequately predicted the suitable objects for most applications. We also appended the class name `object' for all datasets. The final `unknown object' score is the max score for all the unknown object proposals. 
%This can be understood from the text-conditioned results on RWD, which were relatively low for all datasets besides the Aquatic dataset. 

\paragraph{ImageNet Baseline, \baseIN.} For the ImageNet baseline, we used all the class names from ImageNet as unknown object proposals (after filtering out the known objects). The final `unknown object' score is the maximum score of all the ImageNet classes.

\begin{table*}[t]
    \centering
  \begin{tabular}{c|p{6cm}|p{5.2cm}}
    \toprule
    Dataset & Unknown Objects & LLM Proposed Unknowns\\
    \midrule
    Aquatic & \textbf{Puffin}, \textbf{Stingray}, \textbf{Starfish}. & \textbf{Puffin}, \textbf{Stingray}, \textbf{Starfish}, Squid, Dolphin, Lobster, Crab, Seagull... \\
    
\vspace{-0.1in}&\vspace{-0.1in}&\vspace{-0.1in}\\
    
    Aerial & Expressway-Service-Area, Expressway-Toll-Station, \textbf{Baseball Field}, \textbf{Bridge}, Chimney, Harbor, Overpass, Tennis Court, \textbf{Train Station}, Windmill. & \textbf{Baseball field}, \textbf{Bridge}, \textbf{Train Station}, Power Plant, Road, Forest, Parking Lots, Lakes, Mountains, Wind Turbines...\\

\vspace{-0.1in}&\vspace{-0.1in}&\vspace{-0.1in}\\
    
    Game & Talon, Lux, Seraphine, Jhin, Taric... &Ahri, Anivia, Annie, Ashe, Brand...  \\
    \vspace{-0.1in}&\vspace{-0.1in}&\vspace{-0.1in}\\

    Medical & Soft Tissue Calcination, \textbf{Ulna}, Fourth Metacarpal Bone, Artefact, First Metacarpal Bone, \textbf{Radius}. & \textbf{Ulna}, \textbf{Radius}, Carpal bones, First metatarsal, Second metatarsal Fingers...
\\

\vspace{-0.1in}&\vspace{-0.1in}&\vspace{-0.1in}\\

    Surgery & Curette, \textbf{Hook}, Scalpel, Bipolar Forceps Down,  \textbf{Scissors}, Pliers Down, Pliers Up. & \textbf{Hook}, \textbf{Scissors}, Forceps, Needle Driver, Clamp, Knife, Medical Instruments...\\
    \bottomrule
  \end{tabular}
    \caption{\textbf{Large Language Model Novel Object Proposals.} The actual unknown objects (Unknown Objects) and the ones proposed by the large language model (LLM Proposed Unknowns) are listed below. The large language model seems able to identify relevant objects in all domains; however, in most of them, it could not identify all. For the Game dataset, which consisted only of character names, the model could not identify \textit{any} base class names.}
      \label{tab:llm_proposal}
\end{table*}

%%%%%%%%%%%%%%%%%%%%%%%%% UPER SECTION %%%%%%%%%%%%%%%%%%%%%%%%%%%
%BASE-B/16
% Task 1
\newcommand{\BURTone}{58.6}
\newcommand{\BCKTone}{61.8}

% Task 2
\newcommand{\BURTtwo}{55.7}
\newcommand{\BPKTtwo}{61.7}
\newcommand{\BCKTtwo}{43.0}
\newcommand{\BBOTtwo}{52.3}

% Task 3
\newcommand{\BURTthree}{54.9}
\newcommand{\BPKTthree}{52.3}
\newcommand{\BCKTthree}{43.0}
\newcommand{\BBOTthree}{48.4}

% Task 4
\newcommand{\BPKTfour}{48.3}
\newcommand{\BCKTfour}{29.2}
\newcommand{\BBOTfour}{43.5}

%BASE-L/16
% Task 1
\newcommand{\LURTone}{69.0}
\newcommand{\LCKTone}{65.7}

% Task 2
\newcommand{\LURTtwo}{66.7}
\newcommand{\LPKTtwo}{65.7}
\newcommand{\LCKTtwo}{52.3}
\newcommand{\LBOTtwo}{59.0}

% Task 3
\newcommand{\LURTthree}{66}
\newcommand{\LPKTthree}{59.0}
\newcommand{\LCKTthree}{53.7}
\newcommand{\LBOTthree}{57.2}

% Task 4
\newcommand{\LPKTfour}{57.2}
\newcommand{\LCKTfour}{50.5}
\newcommand{\LBOTfour}{55.5}

%%%%%%%%%%%%%%%%%%%%%%%%% LOWER SECTION %%%%%%%%%%%%%%%%%%%%%%%%%%%

%%%%%%%%%%%%%%%%%%%%%%%%% UPER SECTION %%%%%%%%%%%%%%%%%%%%%%%%%%%
%BASE-B/16
% Task 1
\newcommand{\LBURTone}{60.5}
\newcommand{\LBCKTone}{59.5}

% Task 2
\newcommand{\LBURTtwo}{58.1}
\newcommand{\LBPKTtwo}{59.4}
\newcommand{\LBCKTtwo}{37.3}
\newcommand{\LBBOTtwo}{47.8}

% Task 3
\newcommand{\LBURTthree}{57.3}
\newcommand{\LBPKTthree}{47.7}
\newcommand{\LBCKTthree}{40.4}
\newcommand{\LBBOTthree}{45.3}

% Task 4
\newcommand{\LBPKTfour}{46.4}
\newcommand{\LBCKTfour}{34.6}
\newcommand{\LBBOTfour}{43.5}

%BASE-L/16
% Task 1
\newcommand{\LLURTone}{69.3}
\newcommand{\LLCKTone}{69.1}

% Task 2
\newcommand{\LLURTtwo}{69.1}
\newcommand{\LLPKTtwo}{69.2}
\newcommand{\LLCKTtwo}{45.2}
\newcommand{\LLBOTtwo}{56.6}

% Task 3
\newcommand{\LLURTthree}{66.6}
\newcommand{\LLPKTthree}{56.5}
\newcommand{\LLCKTthree}{53.4}
\newcommand{\LLBOTthree}{56.6}

% Task 4
\newcommand{\LLPKTfour}{56.9}
\newcommand{\LLCKTfour}{54.3}
\newcommand{\LLBOTfour}{56.3}

\begin{table*}[t]
\centering

\setlength{\tabcolsep}{6pt}
\adjustbox{width=\textwidth}{
\begin{tabular}{@{}l|cc|cccc|cccc|ccc@{}}
\toprule
 \textbf{Task IDs} ($\rightarrow$)
 & \multicolumn{2}{c|}{\textbf{Task 1}} 
 & \multicolumn{4}{c|}{\textbf{Task 2}} 
 & \multicolumn{4}{c|}{\textbf{Task 3}} 
 & \multicolumn{3}{c}{\textbf{Task 4}} \\ \midrule
 
& \cellcolor[HTML]{FFFFED}{U-Recall}&  \multicolumn{1}{c|}{\cellcolor[HTML]{EDF6FF}{mAP ($\uparrow$)}} & \cellcolor[HTML]{FFFFED}{U-Recall} & \multicolumn{3}{c|}{\cellcolor[HTML]{EDF6FF}{mAP ($\uparrow$)}} & \cellcolor[HTML]{FFFFED}{U-Recall} &  \multicolumn{3}{c|}{\cellcolor[HTML]{EDF6FF}{mAP ($\uparrow$)}} & \multicolumn{3}{c}{\cellcolor[HTML]{EDF6FF}{mAP ($\uparrow$)}}  \\

 & \cellcolor[HTML]{FFFFED}($\uparrow$) 
 &  \begin{tabular}[c]{@{}c}CK\end{tabular} 
 & \cellcolor[HTML]{FFFFED}($\uparrow$)
 &  \begin{tabular}[c]{@{}c@{}}PK\end{tabular} 
 & \begin{tabular}[c]{@{}c@{}}CK\end{tabular} 
 & Both 
 & \cellcolor[HTML]{FFFFED}($\uparrow$) 
 & \begin{tabular}[c]{@{}c@{}}PK\end{tabular} 
 & \begin{tabular}[c]{@{}c@{}}CK\end{tabular} 
 & Both & \begin{tabular}[c]{@{}c@{}}PK\end{tabular} 
 & \begin{tabular}[c]{@{}c@{}}CK\end{tabular} & Both \\ \midrule

OW-DETR~\citep{OWDETR} & \cellcolor[HTML]{FFFFED}7.5 & 59.2 & \cellcolor[HTML]{FFFFED}6.2 &  53.6 & 33.5 & 42.9 & \cellcolor[HTML]{FFFFED}5.7& 38.3 & 15.8 & 30.8 & 31.4 & 17.1 & 27.8 \\

CAT~\citep{CAT}  & \cellcolor[HTML]{FFFFED} 23.7 &   60.0 & \cellcolor[HTML]{FFFFED} 19.1&  55.5 & 32.2 & 44.1  & \cellcolor[HTML]{FFFFED} 24.4& 42.8 &18.7 &  34.8 & 34.4 & 16.6 & 29.9 \\

PROB~\citep{PROB}  & \cellcolor[HTML]{FFFFED} 19.4 &   59.5 & \cellcolor[HTML]{FFFFED} 17.4& 55.7 & 32.2 & 44.0  & \cellcolor[HTML]{FFFFED} 19.6 & 43.0 & 22.2 & 36.0 & 35.7 & 18.9 & 31.5 \\

Hyp-OW~\citep{doan2023hyp}  & \cellcolor[HTML]{FFFFED} 23.5  &   59.4  & \cellcolor[HTML]{FFFFED}  20.6 &- & - & 44.0 &  \cellcolor[HTML]{FFFFED} 26.3 &  - &  - &  36.8  & -  &- & 33.6 \\

MEPU-SS~\citep{fang2023unsupervised}  & \cellcolor[HTML]{FFFFED} 30.3 &   60.0  & \cellcolor[HTML]{FFFFED}  30.6 &  57.0 &  33.1 & 44.5 & \cellcolor[HTML]{FFFFED} 30.0 &42.2 & 20.5 &  35.0 & 34.3 & 18.9 & 30.4 \\

\midrule

MEPU-FS~\citep{fang2023unsupervised}  & \cellcolor[HTML]{FFFFED} 31.6 &  60.2  & \cellcolor[HTML]{FFFFED}  30.9 & 57.3 &  33.3 & 44.8 & \cellcolor[HTML]{FFFFED} 30.1 &  42.6 & 21.0 &  35.4 & 34.8 & 18.9 & 30.4 \\

MAVL~\citep{MAVL}  & \cellcolor[HTML]{FFFFED} 50.1 &  64.0  & \cellcolor[HTML]{FFFFED} 49.5 &61.6 & 30.8 & 46.2  & \cellcolor[HTML]{FFFFED} 50.9 &43.8 & 22.7 & 36.8  & 36.2&  20.6 & 32.3\\

\base-B/16   
& \cellcolor[HTML]{FFFFED}  \BURTone 
&  \BCKTone 
& \cellcolor[HTML]{FFFFED} \BURTtwo
&  \BPKTtwo
& \BCKTtwo
& \BBOTtwo  
& \cellcolor[HTML]{FFFFED} \BURTthree 
&    \BPKTthree
& \BCKTthree
& \BBOTthree
&\BPKTfour
& \BCKTfour
&\BBOTfour\\

\baseIN-B/16  & \cellcolor[HTML]{FFFFED}   63.3 & 61.5 & \cellcolor[HTML]{FFFFED}  61.0 & 61.5 & 43.1 & 52.3 &\cellcolor[HTML]{FFFFED}  58.1 & 52.3 & 40.4 & 48.3 & 48.7 & 29.8 & 44.0 \\

\baseLLM-B/16   
& \cellcolor[HTML]{FFFFED}   66.9 & 62.1 & \cellcolor[HTML]{FFFFED} 59.7 & 61.6 & 43.1 & 52.3 & \cellcolor[HTML]{FFFFED} 59.9 & 52.4 & 40.5 & 48.4 & 48.7 & 29.8 & 44.0 

\\

\base-L/14
& \cellcolor[HTML]{FFFFED}   \LURTone 
&   \textbf{\LCKTone }
& \cellcolor[HTML]{FFFFED}  \LURTtwo 
&    \textbf{\LPKTtwo }
& \textbf{\LCKTtwo}
& \textbf{\LBOTtwo }
& \cellcolor[HTML]{FFFFED}  \LURTthree
&  \textbf{\LPKTthree}
& \LCKTthree
& \textbf{\LBOTthree}
& \LPKTfour
& \LCKTfour
& \LBOTfour \\

%%%%%%%%%%%%

\baseIN-L/14 &  \cellcolor[HTML]{FFFFED}   74.8 & 65.4 & \cellcolor[HTML]{FFFFED} \textbf{73.6} & 65.4 & 52.2 & 58.8 & \cellcolor[HTML]{FFFFED}  73.6 & 58.8 & \textbf{53.8} & 57.1 & \textbf{57.5} & \textbf{50.8} & \textbf{55.8} \\
%%%%%%%%%%%%

\baseLLM-L/14
& \cellcolor[HTML]{FFFFED} \textbf{79.0} & \textbf{65.7} &\cellcolor[HTML]{FFFFED}  71.7 & 65.6 & 52.2 & 58.9 & \cellcolor[HTML]{FFFFED} \textbf{78.5}& 58.9 & 53.7 & 57.1 & \textbf{57.5} & \textbf{50.8} &\textbf{ 55.8} \\

\midrule

\baseGT-B/16$^\dagger$  &  \cellcolor[gray]{0.9}71.5 & \cellcolor[gray]{0.95} 62.1 & \cellcolor[gray]{0.95} 68.3 &\cellcolor[gray]{0.95}  62.1 & \cellcolor[gray]{0.95} 43.3 & \cellcolor[gray]{0.95} 52.7 & \cellcolor[gray]{0.95} 64.5 & \cellcolor[gray]{0.95} 52.7 & \cellcolor[gray]{0.95} 40.8 & \cellcolor[gray]{0.95} 48.7 &\cellcolor[gray]{0.95}  48.7 & \cellcolor[gray]{0.95} 29.8 & \cellcolor[gray]{0.95}  44.0 

\\ 

\baseGT-L/14$^\dagger$ & \cellcolor[gray]{0.9} 85.7 
& \cellcolor[gray]{0.95} 65.8 
& \cellcolor[gray]{0.95} 84.6 
& \cellcolor[gray]{0.95} 65.8 
& \cellcolor[gray]{0.95} 52.3 
& \cellcolor[gray]{0.95}  59.1 
& \cellcolor[gray]{0.95}  83.2 
& \cellcolor[gray]{0.95}  59.1 
& \cellcolor[gray]{0.95}   54.3 
&\cellcolor[gray]{0.95}   57.5 
& \cellcolor[gray]{0.95}  57.5 
& \cellcolor[gray]{0.95}  50.8 
& \cellcolor[gray]{0.95}  55.8 \\

\midrule
\midrule

OW-DETR~\citep{OWDETR} & \cellcolor[HTML]{FFFFED}5.7  & 71.5 & \cellcolor[HTML]{FFFFED}6.2 & 62.8 & 27.5 & 43.8 & \cellcolor[HTML]{FFFFED}6.9 & 45.2 & 24.9 & 38.5 & 38.2 & 28.1 & 33.1 \\

CAT~\citep{CAT}  & \cellcolor[HTML]{FFFFED} 24.0  &74.2& \cellcolor[HTML]{FFFFED} 23.0 & 67.6 & 35.5 & 50.7  & \cellcolor[HTML]{FFFFED} 24.6 &51.2 & 32.6 &  45.0 & 45.4 & 35.1 & 42.8 \\

PROB~\citep{PROB} & \cellcolor[HTML]{FFFFED}17.6  & 73.4 & \cellcolor[HTML]{FFFFED} 22.3 &  66.3 & 36.0 & 50.4 & \cellcolor[HTML]{FFFFED} 24.8 & 47.8 & 30.4 & 42.0 & 42.6 & 31.7 & 39.9 \\

Hyp-OW~\citep{doan2023hyp}  & \cellcolor[HTML]{FFFFED} 23.9  & 72.7  & \cellcolor[HTML]{FFFFED}  23.3  &- & - & 50.6  &  \cellcolor[HTML]{FFFFED} 25.4 & - &  - &  46.2  & -  &- & 44.8  \\ 

MEPU-SS~\citep{fang2023unsupervised}  & \cellcolor[HTML]{FFFFED} 33.3 & 74.2 & \cellcolor[HTML]{FFFFED}  34.2 & 67.5 &  41.0 & 53.6 & \cellcolor[HTML]{FFFFED} 33.6 &  50.0 & 37.5 &  45.8 & 43.2 & 33.5 & 40.8 \\

\midrule
MEPU-FS~\citep{fang2023unsupervised}  & \cellcolor[HTML]{FFFFED} 37.9 &  \textbf{74.3}  & \cellcolor[HTML]{FFFFED}  35.8 & 68.0 &  41.9 & 54.3 & \cellcolor[HTML]{FFFFED} 35.7 & 50.2 & 38.3  &  46.2 & 43.7 & 33.7 & 41.2 \\

\base-B/16   
& \cellcolor[HTML]{FFFFED}  \LBURTone 
&  \LBCKTone 
& \cellcolor[HTML]{FFFFED} \LBURTtwo
& \LBPKTtwo
& \LBCKTtwo
& \LBBOTtwo  
& \cellcolor[HTML]{FFFFED} \LBURTthree 
&    \LBPKTthree
& \LBCKTthree
& \LBBOTthree
&\LBPKTfour
& \LBCKTfour
&\LBBOTfour\\

\baseIN-B/16  & \cellcolor[HTML]{FFFFED}    63.6 & 59.2 & \cellcolor[HTML]{FFFFED} 62.6 & 59.0 & 37.3 & 47.6 & \cellcolor[HTML]{FFFFED}61.1 & 47.6 & 40.5 & 45.2  & 46.8 & 35.4 & 44.0 \\

\baseLLM-B/16   
& \cellcolor[HTML]{FFFFED}  48.9 & 59.6 &\cellcolor[HTML]{FFFFED}  65.1 & 59.4 & 37.3 & 47.8 &\cellcolor[HTML]{FFFFED}  48.6 & 47.7 & 40.5 & 45.3 & 46.8 & 35.4 & 44.0 \\

\base-L/14
& \cellcolor[HTML]{FFFFED}   \LLURTone 
&   \LLCKTone 
& \cellcolor[HTML]{FFFFED}  \LLURTtwo 
&   \textbf{ \LLPKTtwo }
& \textbf{\LLCKTtwo}
& \textbf{ \LLBOTtwo  }
& \cellcolor[HTML]{FFFFED}   \LLURTthree
&  \LLPKTthree
& \LLCKTthree
& \textbf{\LLBOTthree}
& \LLPKTfour
& \LLCKTfour
& \LLBOTfour \\

%%%%%%%%%%%%

\baseIN-L/14 &  \cellcolor[HTML]{FFFFED}  \textbf{75.6} & 69.0 &\cellcolor[HTML]{FFFFED}  76.1 & 68.9 & 45.0 & 56.4 & \cellcolor[HTML]{FFFFED}  \textbf{77.4}& 56.3 & 53.4 & 55.4 &  \textbf{57.1}&  \textbf{54.8} &  \textbf{56.5} \\
%%%%%%%%%%%%

\baseLLM-L/14
& \cellcolor[HTML]{FFFFED} 58.9 & 69.1 &\cellcolor[HTML]{FFFFED}   \textbf{80.5}& 69.1 &  \textbf{45.2} & 56.5 &\cellcolor[HTML]{FFFFED}  65.1 &  \textbf{56.5} &  \textbf{53.5} & 55.5 &  \textbf{57.1} &  \textbf{54.8} &  \textbf{56.5} \\

\midrule

\baseGT-B/16$^\dagger$   & \cellcolor[gray]{0.9}  72.5 & \cellcolor[gray]{0.95} 59.6 & \cellcolor[gray]{0.95} 70.0 & \cellcolor[gray]{0.95} 59.7 & \cellcolor[gray]{0.95}37.5 & \cellcolor[gray]{0.95} 48.0 & \cellcolor[gray]{0.95} 64.5 & \cellcolor[gray]{0.95} 48.0 & \cellcolor[gray]{0.95}40.7 & \cellcolor[gray]{0.95} 45.6 & \cellcolor[gray]{0.95}46.8 & \cellcolor[gray]{0.95} 35.4 & \cellcolor[gray]{0.95}44.0 \\

\baseGT-L/14$^\dagger$ &\cellcolor[gray]{0.9} 85.9 & \cellcolor[gray]{0.95}69.1 &\cellcolor[gray]{0.95} 85.4 &\cellcolor[gray]{0.95} 69.2 &\cellcolor[gray]{0.95} 45.2 & \cellcolor[gray]{0.95}56.6 & \cellcolor[gray]{0.95}83.0 &\cellcolor[gray]{0.95} 56.6 &\cellcolor[gray]{0.95} 54.0 & \cellcolor[gray]{0.95}55.7 &\cellcolor[gray]{0.95} 57.1 &\cellcolor[gray]{0.95} 54.8 & \cellcolor[gray]{0.95}56.5 \\

\bottomrule
\end{tabular}}
\caption{\textbf{Baseline results on M-(top) and S-(bottom) OWODB.} Comparison of unknown class recall and mAP (U-R/U-mAP), and (previously/currently) known class mAP@$0.5$ of $0$-shot \base\ to previous OWD methods. 
As there is no domain shift, few-shot adaptation is not required to get adequate results. Only MAVL, MEPU-FS, and \base\ foundation models.}
\label{table:t1_owod}
\end{table*}

\begin{table*}[t]
\centering

\setlength{\tabcolsep}{10pt}
\adjustbox{width=\textwidth}{
\begin{tabular}{@{}l|ccc|ccc|ccc}
\toprule
 \textbf{Task IDs} ($\rightarrow$)& \multicolumn{3}{c|}{\textbf{Task 1}} & \multicolumn{3}{c|}{\textbf{Task 2}} & \multicolumn{3}{c}{\textbf{Task 3}} \\
\midrule

 & \cellcolor[HTML]{FFFFED}{U-Recall} & \cellcolor[HTML]{EDF6FF}{WI} & \cellcolor[HTML]{EDF6FF}{A-OSE} & \cellcolor[HTML]{FFFFED}{U-Recall} & \cellcolor[HTML]{EDF6FF}{WI} & \cellcolor[HTML]{EDF6FF}{A-OSE}  & \cellcolor[HTML]{FFFFED}{U-Recall} & \cellcolor[HTML]{EDF6FF}{WI} & \cellcolor[HTML]{EDF6FF}{A-OSE} \\

 & \cellcolor[HTML]{FFFFED}($\uparrow$) & \cellcolor[HTML]{EDF6FF}($\downarrow$) & \cellcolor[HTML]{EDF6FF}($\downarrow$) & \cellcolor[HTML]{FFFFED}($\uparrow$) & \cellcolor[HTML]{EDF6FF}($\downarrow$) & \cellcolor[HTML]{EDF6FF}($\downarrow$) & \cellcolor[HTML]{FFFFED}($\uparrow$) & \cellcolor[HTML]{EDF6FF}($\downarrow$) & \cellcolor[HTML]{EDF6FF}($\downarrow$) \\

 \midrule

ORE $-$ EBUI ~\cite{TowardsOWOD} & \cellcolor[HTML]{FFFFED}4.9  & 0.0621 & 10459 & \cellcolor[HTML]{FFFFED}2.9 & 0.0282 & 10445 & \cellcolor[HTML]{FFFFED}3.9 & 0.0211 & 7990  \\

2B-OCD ~\cite{two_branch_OWOD} & \cellcolor[HTML]{FFFFED}12.1 & 0.0481 & - & \cellcolor[HTML]{FFFFED}9.4 & 0.160 & - & \cellcolor[HTML]{FFFFED}11.6 & \textbf{ 0.0137} & -  \\

OW-DETR \cite{OWDETR} & \cellcolor[HTML]{FFFFED}7.5  & 0.0571 & 10240 & \cellcolor[HTML]{FFFFED}6.2 & 0.0278 & 8441 & \cellcolor[HTML]{FFFFED}5.7 & 0.0156 & 6803  \\

OCPL ~\cite{OWOD_OCPL} & \cellcolor[HTML]{FFFFED} 8.3  & \textbf{0.0413} & 5670 & \cellcolor[HTML]{FFFFED} 7.6 & \textbf{ 0.0220 }& \textbf{ 5690 }& \cellcolor[HTML]{FFFFED}11.9 & 0.0162 & 5166  \\

PROB~\cite{PROB} & \cellcolor[HTML]{FFFFED} \textbf{19.4}  & 0.0569 & \textbf{ 5195} & \cellcolor[HTML]{FFFFED} \textbf{ 17.4 }& 0.0344 & 6452 & \cellcolor[HTML]{FFFFED} \textbf{ 19.6 }& 0.0151 & \textbf{ 2641 } \\

\midrule
\base-B/16   & \cellcolor[HTML]{FFFFED} 58.6 & \textbf{ 0.0320 }&\textbf{  6731} & \cellcolor[HTML]{FFFFED} 55.7 & 0.0225 & \textbf{ 6524} & \cellcolor[HTML]{FFFFED} 54.9&\textbf{  0.0130 }&\textbf{4162}  \\

\baseIN-B/16  & \cellcolor[HTML]{FFFFED}     63.3 & 0.0322 & 9002 &\cellcolor[HTML]{FFFFED} 61.0 & 0.0214 & 8441 &\cellcolor[HTML]{FFFFED} 58.1 & 0.0148 & 5960\\

\baseLLM-B/16   
& \cellcolor[HTML]{FFFFED}   66.9 & 0.0343 & 14128 &\cellcolor[HTML]{FFFFED} 59.7 & 0.0243 & 10789 & \cellcolor[HTML]{FFFFED}59.9 & 0.0154 & 6702 \\

%%%%%%%%%%%%

\base-L/14 & \cellcolor[HTML]{FFFFED} 69.0 & 0.0337& 8796 & \cellcolor[HTML]{FFFFED} 66.7&  0.0211 & 8628 & \cellcolor[HTML]{FFFFED} 66.0 & 0.0137& 5730  \\

\baseIN-L/14 & \cellcolor[HTML]{FFFFED} 74.8 & 0.0337 & 8835& \cellcolor[HTML]{FFFFED}  \textbf{73.6} & 0.0201 & 8043& \cellcolor[HTML]{FFFFED} 73.6 & 0.0131 & 5864 \\

%%%%%%%%%%%%

\baseLLM-L/14
& \cellcolor[HTML]{FFFFED}   \textbf{79.0} & 0.0331 & 15118 &\cellcolor[HTML]{FFFFED} 71.7& \textbf{0.0200}& 9776 &\cellcolor[HTML]{FFFFED} \textbf{78.5} & 0.0131 & 6653\\

\midrule

\baseGT-B/16$^\dagger$  &  \cellcolor[gray]{0.9} 71.5 &
\cellcolor[gray]{0.95} 0.0358  & 
\cellcolor[gray]{0.95} 21743  & 
\cellcolor[gray]{0.95 } 68.3   & 
\cellcolor[gray]{0.95} 0.0212  & 
\cellcolor[gray]{0.95 } 18927  & 
\cellcolor[gray]{0.95} 64.5    &
\cellcolor[gray]{0.95} 0.0149 &
\cellcolor[gray]{0.95} 12919 \\

\baseGT-L/14$^\dagger$  & \cellcolor[gray]{0.9} 85.7 &  \cellcolor[gray]{0.95} 0.0285 &  \cellcolor[gray]{0.95} 20563 &  \cellcolor[gray]{0.95} 84.6 &  \cellcolor[gray]{0.95} 0.0161 & \cellcolor[gray]{0.95}  17227 & \cellcolor[gray]{0.95}  83.2 &  \cellcolor[gray]{0.95} 0.0116 & \cellcolor[gray]{0.95} 12432 \\
\bottomrule

\end{tabular}%
}
\caption{ 
\textbf{Unknown Object Confusion on M-OWODB.} The comparison is shown in terms of wilderness impact (WI), absolute open set error (A-OSE) and unknown class recall (U-Recall). The unknown recall (U-Recall) metric quantifies a model's ability to retrieve the unknown object instances. 
\ourMethod\ achieves improved WI, A-OSE and U-Recall over OW-DETR across tasks, thereby indicating less confusion in detecting unknown instances as known classes with higher unknown instance detection capabilities. See Sec.~\ref{sup:sec:owod} for additional details.
}
\label{tab:wi_ose}

\end{table*}

\paragraph{Attribute Adaptation.} 
There exists a gap  between vision and text-derived embeddings~\cite{ModalityGap}. While the visual and text gap is orthogonal, this makes refining text-derived embeddings from visual embeddings problematic. Therefore, we adapt the the selected attributes based on the extracted visual embeddings before performing attribute refinement. When running an $N_s$ shot experiment, we are given $N_s$ examples per class. 
Specifically, for each class $c \in C$:
\begin{equation}
    \mathbf{E}_{cls}^v[c] = \frac{1}
    {N_s}\sum_{i=0}^{N_s}\mathbf{e}^v_i,~~\mathbf{e}^v_i\in C.
\end{equation}
We then use these average vision-derived class embeddings to adapt the text-derived attributes.  We found that refining without first performing such an alignment results in poor performance - as we are limited in the amount of supervision we have (with more supervision, such alignment is unnecessary). To perform alignment, we optimize $\mathbf{E}_{att}$ using the $L_2$ loss, while keeping $\mathbf{W}$ frozen:
\begin{equation}
    ||\mathbf{WE}_{att} - \mathbf{E}_{cls}^v ||_2.
\end{equation}

\paragraph{Prompt Ensembling.}
We utilized "prompt templates" in OWL-ViT, defined as simple text patterns like "a photo of a \{\}". The placeholder is replaced with the category name. These templates help bridge the gap between image-level pre-training and detection fine-tuning. For ensembling, multiple templates are used during inference. Specifically, the model is tested against a set of the "7 best" CLIP prompts, as described in~\cite{owl-vit}. The text embeddings from these templates are averaged per class to produce the final text-derived class embedding.

\paragraph{Computation.}
All experiments were performed on 2 Nvidia A6000 systems, except the ImageNet baselines, which required more memory (due to the large number of classes, ImageNet baselines require systems with more GPU memory), and one A100 80GB was used.

\section{Additional Results}
In this section, we provide additional quantitative and qualitative results. Specifically, in Sec.~\ref{sup:sec:owod}, we provide all the standard OWD benchmark results, from the popular S- and M- OWODB results to the reporting of the unknown confusion metrics (WI and AOSE) on M-OWODB. In Sec.~\ref{sup:sec:fs_exp}, we show the effect of varying the number of shots on \ourMethod\ and \baseFS's performance. In Sec.~\ref{sup:sec:zs_ood}, we discuss how the zero-shot results on RWD indicate the out-of-distribution nature of the dataset. In Sec.~\ref{sup:sec:full_ablation}, we ablate \ourMethod's various components with a per-dataset breakdown. In Sec.~\ref{sup:sec:attribute_study}, we study the relationship of attributes to the known and unknown classes. Lastly, in Sec.~\ref{sup:sec:more_qual}, we provide additional qualitative results, including failure cases. 

\subsection{Results on Existing Benchmarks}
\label{sup:sec:owod}

\paragraph{Full Open World Object Detection Benchmark Results.} In Tab.~\ref{table:t1_owod}, we present the full results of our baseline models on the S- and M-OWODB benchmarks. These benchmarks provide distinct evaluation conditions: S-OWODB focuses on clear super-category separation, while M-OWODB introduces classes in a mixed, progressive manner. Our results demonstrate that all baselines achieve remarkable performance on both benchmarks. In the M-OWODB, \baseLLM\ surpasses \baseIN, a trend that reverses in the S-OWODB. This difference likely arises from the super-category separation in S-OWODB, which complicates the accurate prediction of unknown classes for LLMs.

The small performance differential between \baseLLM\ and \baseGT\ across both benchmarks indicates a saturation point in the potential for improvement using these methodologies. This is particularly evident with the high performance of \baseIN, suggesting that open-vocabulary detection methods are close to resolving the challenges posed by these benchmarks. On M-OWODB, \baseLLM\ consistently maintains a U-Recall above $71.7$ and a known mAP above $55.8$ across all tasks. On S-OWODB, \baseIN\ achieves a U-Recall exceeding $75.6$ and a known mAP of 69.0 across all tasks. These findings suggest that the effectiveness of open-vocabulary detection methods in these benchmarks is nearing its peak. Consequently, the use of S- and M-OWODB for evaluating open world detection methods utilizing foundation models might no longer be as challenging or informative, pointing to the need for more advanced and demanding benchmarks to better assess and drive the future development of OWD methodologies.

For a discussion on why our baselines are not directly comparable to existing OWD methods, please see Sec.~\ref{sec:limitations}.

\begin{table*}[t]
\centering
\setlength{\tabcolsep}{3pt}
\adjustbox{width=\textwidth}{
\Large
\begin{tabular}{@{}ll|cc|cc|cc|cc|cc|cc|cc|cc|cc|cc||cc|cc@{}}
\toprule
 &\textbf{Domain} & \multicolumn{4}{c|}{\textbf{Aquatic}} & \multicolumn{4}{c|}{\textbf{Aerial}} & \multicolumn{4}{c|}{\textbf{Game}}& \multicolumn{4}{c|}{\textbf{Medical}} & \multicolumn{4}{c||}{\textbf{Surgery}}&  \multicolumn{4}{c}{\cellcolor[HTML]{EDF6FF} \textbf{Overall}} \\ 
 
 \midrule
 \multicolumn{2}{c|}{\textbf{Task}}& \multicolumn{2}{c|}{\textbf{Task 1}} & \multicolumn{2}{c|}{\textbf{Task 2}} & \multicolumn{2}{c|}{\textbf{Task 1}} & \multicolumn{2}{c|}{\textbf{Task 2}} 
& \multicolumn{2}{c|}{\textbf{Task 1}} & \multicolumn{2}{c|}{\textbf{Task 2}} 
& \multicolumn{2}{c|}{\textbf{Task 1}} & \multicolumn{2}{c|}{\textbf{Task 2}} 
& \multicolumn{2}{c|}{\textbf{Task 1}} 
& \multicolumn{2}{c||}{\textbf{Task 2}}
& \multicolumn{2}{c|}{\cellcolor[HTML]{EDF6FF}\textbf{Task 1}} & \multicolumn{2}{c}{\cellcolor[HTML]{EDF6FF}\textbf{Task 2}}\\

  &&
  \cellcolor[HTML]{FFFFED} U & K & PK & CK &
  \cellcolor[HTML]{FFFFED} U & K & PK & CK &
  \cellcolor[HTML]{FFFFED} U & K & PK & CK & 
  \cellcolor[HTML]{FFFFED} U & K & PK & CK & 
  \cellcolor[HTML]{FFFFED} U & K & PK & CK & 
  \cellcolor[HTML]{FFFFED} U & K & PK & CK \\

\midrule

\multirow{4}{*}{\rotatebox[origin=c]{90}{$1$-shot}}

 & \base-B/16& \cellcolor[HTML]{FFFFED} 7.1 & 25.7 & 22.8 & 20.4 
 & \cellcolor[HTML]{FFFFED} 1.2 & 6.4 & 6.8 & 6.7 
 & \cellcolor[HTML]{FFFFED} 16.0 & 2.1 & 1.6 & 1.4  
 & \cellcolor[HTML]{FFFFED} 0.6 & 3.7 & 3.8 & 3.7   
 & \cellcolor[HTML]{FFFFED} 1.3 & 4.7 & 5.7 & 5.3  
 & \cellcolor[HTML]{FFFFED} 5.2 & 8.5 & 8.1 & 7.5 \\

 & \ourMethod-B/16 
 & \cellcolor[HTML]{FFFFED}  14.2 & 22.2 & 18.4 & 19.9
 & \cellcolor[HTML]{FFFFED}  2.0 & 6.1 & 6.0 & 6.3
 & \cellcolor[HTML]{FFFFED}  5.9 & 1.7 & 1.2 & 1.1
 & \cellcolor[HTML]{FFFFED}  0.8 & 3.9 & 2.9 & 2.8
 & \cellcolor[HTML]{FFFFED}  2.3 & 5.4 & 5.3 & 5.3
 & \cellcolor[HTML]{FFFFED}  5.04 & 7.86 & 6.76 & 7.08 \\

 & \base-L/14& \cellcolor[HTML]{FFFFED} 2.4 & 18.1 & 17.4 & 16.9 
 & \cellcolor[HTML]{FFFFED} 9.7 & 15.8 & 15.9 & 13.2 
 & \cellcolor[HTML]{FFFFED} 8.2 & 9.0 & 8.7 & 5.8 
 & \cellcolor[HTML]{FFFFED} 1.1 & 20.8 & 20.2 & 21.3   
 & \cellcolor[HTML]{FFFFED} 3.6 & 25.0 & 24.2 & 11.1 
 & \cellcolor[HTML]{FFFFED} 5.0 & \textbf{17.7} & \textbf{17.3} & \textbf{13.7} \\

 & \ourMethod-L/14 
 & \cellcolor[HTML]{FFFFED}  18.0 & 18.1 & 17.4 & 17.0
 & \cellcolor[HTML]{FFFFED}  3.1  & 15.6 & 15.7 & 12.7
 & \cellcolor[HTML]{FFFFED}  28.3 & 7.2 & 5.2 & 4.6
 & \cellcolor[HTML]{FFFFED}  6.1  & 20.5 & 14.8 & 22.7
 & \cellcolor[HTML]{FFFFED}  11.5 & 25.0 & 24.3 & 11.4
 & \cellcolor[HTML]{FFFFED}  \textbf{13.4} & 17.2 & 15.5 & \textbf{13.7} \\

\midrule

\multirow{4}{*}{\rotatebox[origin=c]{90}{$10$-shot}}

 & \base-B/16 
 & \cellcolor[HTML]{FFFFED} 7.1 & 37.8 & 37.9 & 28.1   
 & \cellcolor[HTML]{FFFFED} 1.2 & 8.6 & 8.7 & 1.8  
 & \cellcolor[HTML]{FFFFED} 16.0 & 4.1 & 4.2 & 3.1  
 & \cellcolor[HTML]{FFFFED} 0.6 & 5.9 & 5.9 & 1.7   
 & \cellcolor[HTML]{FFFFED} 1.3 & 11.9 & 13.3 & 9.6  
 & \cellcolor[HTML]{FFFFED} 5.2 & 13.7 & 14.0 & 8.9 \\

 & \ourMethod-B/16 
 & \cellcolor[HTML]{FFFFED}  10.0 & 37.4 & 36.2 & 29.8
 & \cellcolor[HTML]{FFFFED}  1.3 & 9.5 & 9.8 & 2.0
 & \cellcolor[HTML]{FFFFED}  11.0 & 3.8 & 4.1 & 3.3
 & \cellcolor[HTML]{FFFFED}  4.8 & 5.9 & 6.0 & 1.2
 & \cellcolor[HTML]{FFFFED}  13.1 & 12.9 & 14.4 & 10.7
 & \cellcolor[HTML]{FFFFED}  8.0 & 13.9 & 14.1 & 9.4 \\

 & \base-L/14
 & \cellcolor[HTML]{FFFFED} 2.4 & 37.0 & 36.5 & 27.6 
 & \cellcolor[HTML]{FFFFED} 9.7 & 21.8  & 21.1& 6.8 
 & \cellcolor[HTML]{FFFFED} 8.2 & 11.4 & 11.2 & 12.7 
 & \cellcolor[HTML]{FFFFED} 1.1 & 27.3 & 25.8 & 27.5 
 & \cellcolor[HTML]{FFFFED} 3.6 & 24.1 & 23.7 & 7.6  
 & \cellcolor[HTML]{FFFFED} 5.0 & 24.3 & 23.7 & 16.4 \\

 & \ourMethod-L/14 
 & \cellcolor[HTML]{FFFFED}  12.8 & 37.2 & 36.5 & 27.6
 & \cellcolor[HTML]{FFFFED}  5.6 & 22.0 & 21.5 & 7.9
 & \cellcolor[HTML]{FFFFED}  30.3 & 11.6 & 10.6 & 11.2
 & \cellcolor[HTML]{FFFFED}  13.6 & 25.4 & 24.0 & 33.0
 & \cellcolor[HTML]{FFFFED}  11.3 & 26.8 & 28.0 & 11.3
 & \cellcolor[HTML]{FFFFED}  \textbf{14.7} & \textbf{24.6} & \textbf{24.1} & \textbf{18.2} \\

 \midrule
 \midrule

\multirow{4}{*}{\rotatebox[origin=c]{90}{$100$-shot}}

 & \base-B/16 
 & \cellcolor[HTML]{FFFFED} 7.1 & 41.1 & 41.1 & 31.9 
 & \cellcolor[HTML]{FFFFED} 1.2 & 10.4 & 10.1 & 4.0 
 & \cellcolor[HTML]{FFFFED} 16.0 & 4.6 & 4.8 & 3.9  
 & \cellcolor[HTML]{FFFFED} 0.6 & 6.1 & 6.1 & 3.3  
 & \cellcolor[HTML]{FFFFED} 1.3 & 11.9 & 11.3 & 10.9 
 & \cellcolor[HTML]{FFFFED} 5.2 & 14.8 & 14.7 & 10.8 \\

 & \ourMethod-B/16 
 & \cellcolor[HTML]{FFFFED} 3.5 & 43.8 & 44.1 & 40.8
 & \cellcolor[HTML]{FFFFED} 0.9 & 12.0 & 12.6 & 5.4 
 & \cellcolor[HTML]{FFFFED} 13.3 & 3.8 & 4.4 & 4.1  
 & \cellcolor[HTML]{FFFFED} 2.1 & 6.4 & 5.5 & 11.5
 & \cellcolor[HTML]{FFFFED} 6.1 & 12.7 & 12.9 & 11.0
 & \cellcolor[HTML]{FFFFED} 5.2 & 15.7 & 15.9 & 14.6 \\

 & \base-L/14 
 & \cellcolor[HTML]{FFFFED} 2.4 & 43.6 & 42.9 & 42.8 
 & \cellcolor[HTML]{FFFFED} 9.7 & 23.7  & 21.9& 13.0 
 & \cellcolor[HTML]{FFFFED} 8.2 & 10.4 & 10.2 & 13.4 
 & \cellcolor[HTML]{FFFFED} 1.1 & 23.2 & 21.7 & 24.2 
 & \cellcolor[HTML]{FFFFED} 3.6 & 26.0 & 25.0 & 7.4  
 & \cellcolor[HTML]{FFFFED} 5.0 & 25.4 & 24.3 & 20.2 \\

 & \ourMethod-L/14 
 & \cellcolor[HTML]{FFFFED}  18.2 & 50.1 & 48.1 & 47.1
 & \cellcolor[HTML]{FFFFED} 6.0 & 25.3 & 23.7 & 16.0 
 & \cellcolor[HTML]{FFFFED} 30.4 & 10.7 & 9.9 & 11.2
 & \cellcolor[HTML]{FFFFED} 9.4 & 21.8 & 19.9 & 34.6 
 & \cellcolor[HTML]{FFFFED} 12.0 & 29.0 & 28.9 & 8.5
 & \cellcolor[HTML]{FFFFED} \textbf{15.2} & \textbf{27.4} & \textbf{26.1} & \textbf{23.5} \\
\bottomrule

\end{tabular}%
}
\caption{\textbf{Few-shot Analysis on the RWD benchmark.} 
Each domain is broken up into two tasks, where in task 1 you are given some classes, and then known and unknown mAP are evaluated. In Task 2, the rest of the classes are relead and we evaluate Prevously and currently known mAP. We report: $1$, $10$ and $100$-shot respectively.}
\label{table:rwod_fs}
\end{table*}

\begin{figure*}[t]
  \centering
  \includegraphics[width=\textwidth]{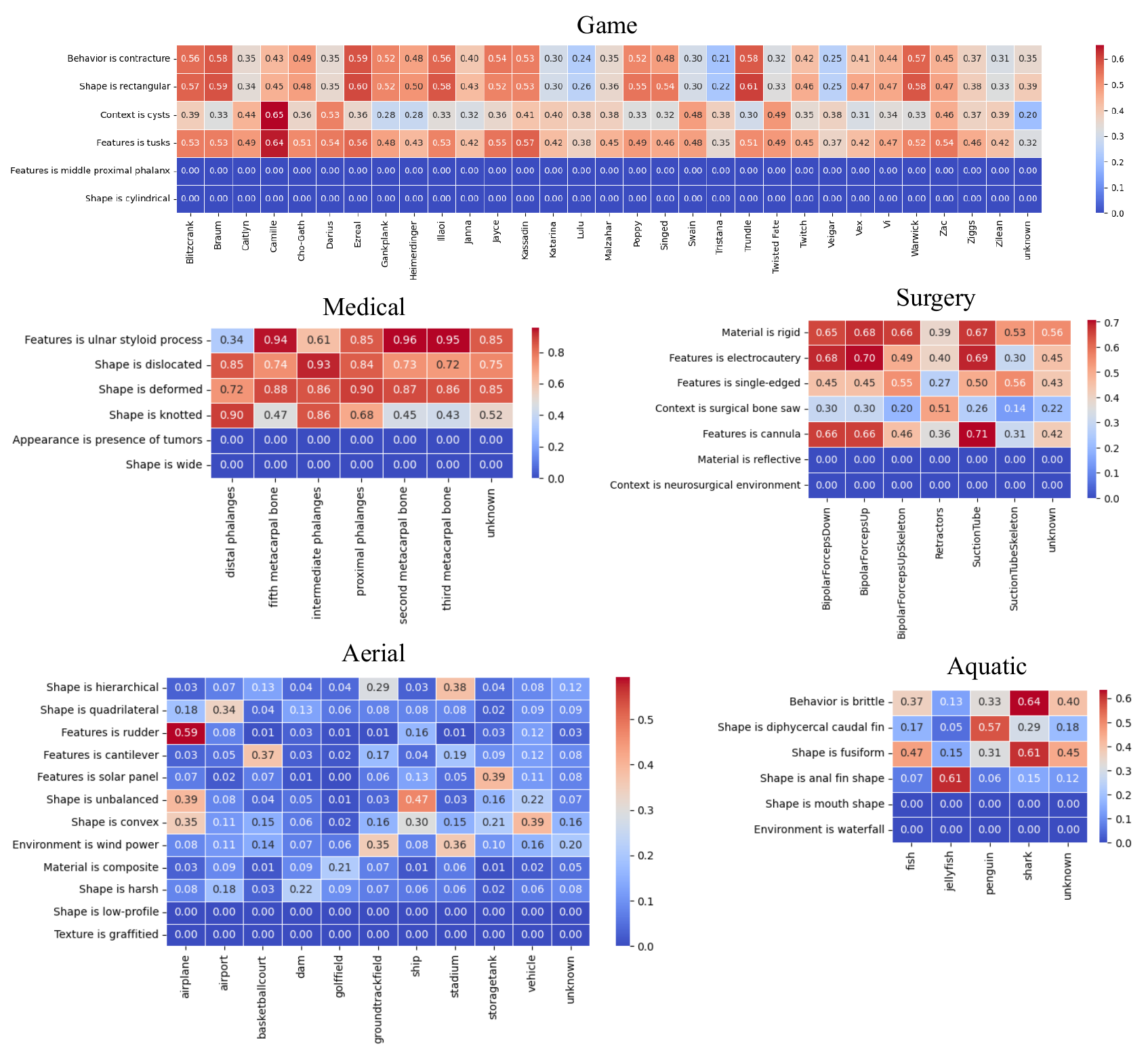}
  \caption{ \textbf{Attribute Study Across Datasets.} This figure displays the average activation scores of attributes for each class in multiple datasets, including Aquatic, Aerial, Game, Medical, and Surgery. The heatmap illustrates the intensity of attribute activation, revealing common and unique attributes across classes. It also highlights the alignment of selected unknown objects with high activation scores, demonstrating the role of attributes in identifying unknown objects.
  }
\label{fig:att_study}
\end{figure*}
\begin{table*}[t]
\centering
\setlength{\tabcolsep}{3pt}
\adjustbox{width=\textwidth}{
\Large
\begin{tabular}{@{}ll|cc|cc|cc|cc|cc|cc|cc|cc|cc|cc||cc|cc@{}}
\toprule
 &\textbf{Domain} & \multicolumn{4}{c|}{\textbf{Aquatic}} & \multicolumn{4}{c|}{\textbf{Aerial}} & \multicolumn{4}{c|}{\textbf{Game}}& \multicolumn{4}{c|}{\textbf{Medical}} & \multicolumn{4}{c||}{\textbf{Surgery}}&  \multicolumn{4}{c}{\cellcolor[HTML]{EDF6FF} \textbf{Overall}} \\ 
 
 \midrule
 \multicolumn{2}{c|}{\textbf{Task}}& \multicolumn{2}{c|}{\textbf{Task 1}} & \multicolumn{2}{c|}{\textbf{Task 2}} & \multicolumn{2}{c|}{\textbf{Task 1}} & \multicolumn{2}{c|}{\textbf{Task 2}} 
& \multicolumn{2}{c|}{\textbf{Task 1}} & \multicolumn{2}{c|}{\textbf{Task 2}} 
& \multicolumn{2}{c|}{\textbf{Task 1}} & \multicolumn{2}{c|}{\textbf{Task 2}} 
& \multicolumn{2}{c|}{\textbf{Task 1}} 
& \multicolumn{2}{c||}{\textbf{Task 2}}
& \multicolumn{2}{c|}{\cellcolor[HTML]{EDF6FF}\textbf{Task 1}} & \multicolumn{2}{c}{\cellcolor[HTML]{EDF6FF}\textbf{Task 2}}\\

  &&
  \cellcolor[HTML]{FFFFED} U & K & PK & CK &
  \cellcolor[HTML]{FFFFED} U & K & PK & CK &
  \cellcolor[HTML]{FFFFED} U & K & PK & CK & 
  \cellcolor[HTML]{FFFFED} U & K & PK & CK & 
  \cellcolor[HTML]{FFFFED} U & K & PK & CK & 
  \cellcolor[HTML]{FFFFED} U & K & PK & CK \\

\midrule

 & FOMO-L/14 
 & \cellcolor[HTML]{FFFFED}  18.2 & 50.1 & 48.1 & 47.1
 & \cellcolor[HTML]{FFFFED} 6.0 & 25.3 & 23.7 & 16.0 
 & \cellcolor[HTML]{FFFFED} 30.4 & 10.7 & 9.9 & 11.2
 & \cellcolor[HTML]{FFFFED} 9.4 & 21.8 & 19.9 & 34.6 
 & \cellcolor[HTML]{FFFFED} 12.0 & 29.0 & 28.9 & 8.5
 & \cellcolor[HTML]{FFFFED} 15.2 & 27.4 & 26.1 & 23.5 \\

 & - Refinement & \cellcolor[HTML]{FFFFED}18.4 & 43.4 &43.1 & 42.6 & \cellcolor[HTML]{FFFFED}4.4 & 23.8 & 22.1 & 12.5 & \cellcolor[HTML]{FFFFED} 30.4 & 9.9 &  9.9 & 12.2 & \cellcolor[HTML]{FFFFED}10.3 & 21.3 & 20.2 & 27.8 &\cellcolor[HTML]{FFFFED} 11.9 & 26.1 &25.8 & 8.3 & \cellcolor[HTML]{FFFFED} 15.1 & 24.9 & 24.2 & 20.7 \\
 
  & - Selection& \cellcolor[HTML]{FFFFED}18.4 & 24.3 &11.4 & 29.7 & \cellcolor[HTML]{FFFFED}6.1 & 2.2 & 1.1 & 1.8 & \cellcolor[HTML]{FFFFED}30.4 & 2.2 &1.0 & 0.9 & \cellcolor[HTML]{FFFFED}11.7 & 1.5 & 0.0 & 0.0 &\cellcolor[HTML]{FFFFED} 11.9 & 17.9 & 3.9 & 2.4 &\cellcolor[HTML]{FFFFED}  15.7 &9.6 & 3.5 & 7.0 \\

  & - Adaption &\cellcolor[HTML]{FFFFED}  0.3 & 0.1 & 0.1 & 0.1 &\cellcolor[HTML]{FFFFED}  0.6 & 0.1 & 0.1 & 0.1 & \cellcolor[HTML]{FFFFED} 1.6 & 0.1 & 0.1 & 0.0 &\cellcolor[HTML]{FFFFED}  0.1 & 0.0 & 0.0 & 0.0 &\cellcolor[HTML]{FFFFED}  1.2 & 1.0 & 0.7 & 0.1 & \cellcolor[HTML]{FFFFED} 0.7 & 0.3 & 0.2 & 0.1  \\

\bottomrule

\end{tabular}%
}
\caption{\textbf{Full Ablation} Each domain is broken up into two tasks, where in task 1 you are given some classes, and then known and unknown mAP are evaluated. In Task 2, the rest of the classes are revlead and we evaluate Prevously and currently known mAP. We report: $1$, $10$ and $100$-shot respectively.}

\label{table:full_ablation}
\end{table*}

\paragraph{AOSE-WI Analysis.} Tab.~\ref{tab:wi_ose} presents a detailed analysis of Absolute Open Set Error (AOSE) and Wilderness Impact (WI) for various models tested on the M-OWODB benchmark. 
For concise definitions of these metrics, see~\cite{PROB}.
These metrics are pivotal in assessing the efficacy of object detection models, especially in distinguishing between known and unknown objects. AOSE measures the frequency of misclassifying unknown objects as known, while WI quantifies the impact of unknown objects on the detection of known objects. Therefore, a lower AOSE and WI indicate a model's superior capability to identify unknown objects and minimize confusion with known classes. However, by definition, both tend to increase with higher unknown recall. Similar to previous methods, our baselines were limited to 100 predictions per image.

\begin{figure*}
  \centering
  \begin{subfigure}{0.19\textwidth}
    \centering
    Aquatic
  \end{subfigure}
  \begin{subfigure}{0.17\textwidth}
    \centering
    Aerial
  \end{subfigure}
  \begin{subfigure}{0.22\textwidth}
    \centering
    Game
  \end{subfigure}
  \begin{subfigure}{0.16\textwidth}
    \centering
    Medical
  \end{subfigure}
  \begin{subfigure}{0.22\textwidth}
      \centering
  Surgery
    \end{subfigure}

    \centering
  \includegraphics[width=1\textwidth]{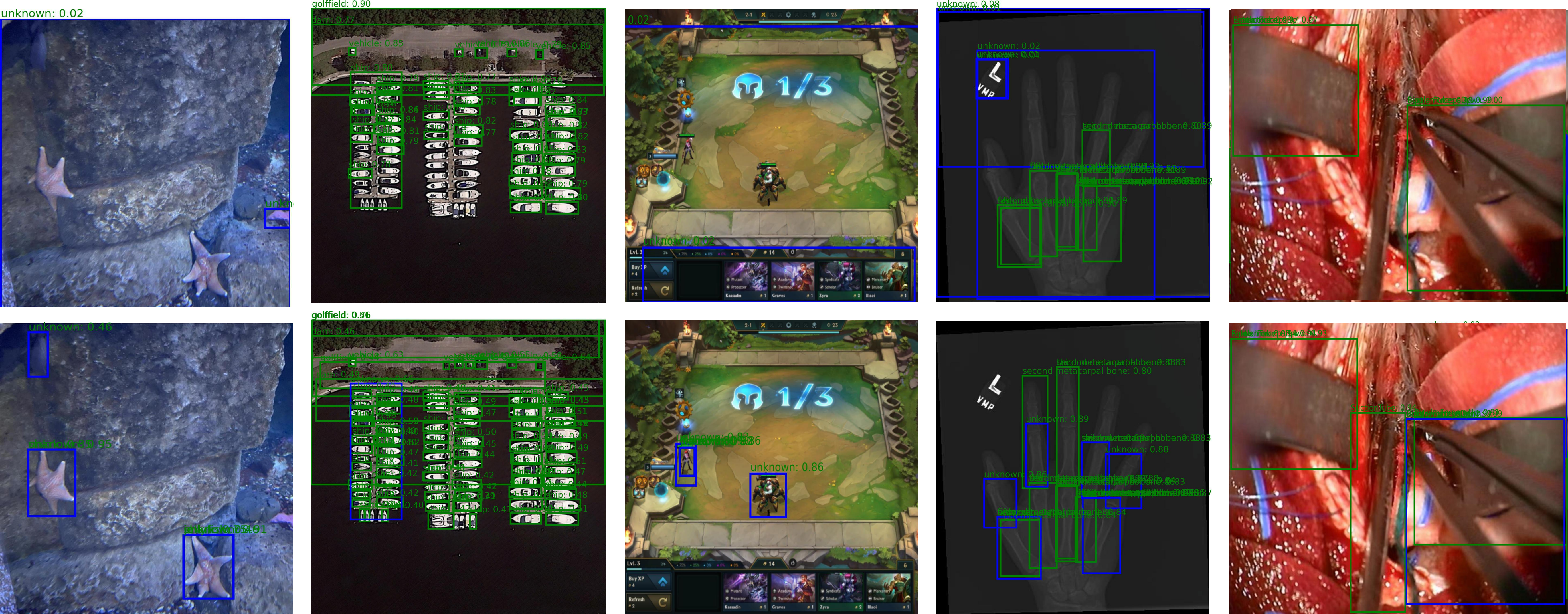}
    \vspace{-0.08in}
  \caption{\textbf{Qualitative Results: Favorable Cases} Qualitative results (top) \baseFS\ (bottom) \ourMethod\ on RWD. {\color{blue}blue} - unknown, {\color{green}green} - known.
  \ourMethod\ shows superior performance on RWD, appearing to have less known-class confusion and better unknown object detection capability. 
  }\vspace{-0.09in}
\label{fig:qualitative_good}
\end{figure*}

\begin{figure*}
  \centering
  \begin{subfigure}{0.19\textwidth}
    \centering
    Aquatic
  \end{subfigure}
  \begin{subfigure}{0.17\textwidth}
    \centering
    Aerial
  \end{subfigure}
  \begin{subfigure}{0.22\textwidth}
    \centering
    Game
  \end{subfigure}
  \begin{subfigure}{0.16\textwidth}
    \centering
    Medical
  \end{subfigure}
  \begin{subfigure}{0.22\textwidth}
      \centering
  Surgery
    \end{subfigure}

    \centering
  \includegraphics[width=1\textwidth]{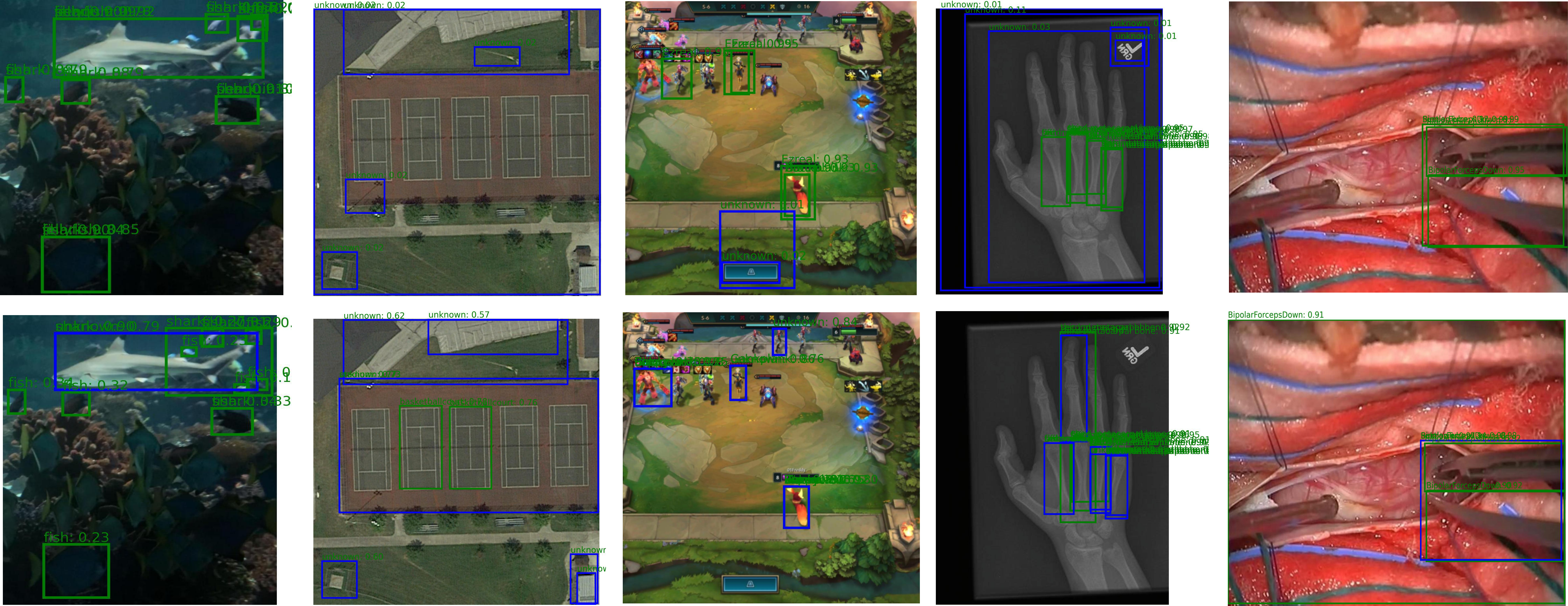}
  \vspace{-0.08in}
  \caption{\textbf{Qualitative Results: Failure Cases} Qualitative results (top) \baseFS\ (bottom) \ourMethod\ on RWD. {\color{blue}blue} - unknown, {\color{green}green} - known.
  \ourMethod\ shows superior performance on RWD, appearing to have less known-class confusion and better unknown object detection capability. 
  }\vspace{-0.09in}
\label{fig:qualitative_bad}
\end{figure*}

\subsection{Effect of number of shots on performance}
\label{sup:sec:fs_exp}
\ourMethod\ is sensitive to the number of shots available for attribute refinement. As seen in Tab.~\ref{table:dataset_details_ood}, many datasets are in the low-data regime. We, therefore, want to see how much \ourMethod\ is sensitive to the number of object exemplars provided.
In Tab.~\ref{table:rwod_fs}, we study the effect of the number of shots on performance.
In the $1$- and $10$- shot experiments, \baseFS\ and \ourMethod's known object mAP are similar, showing that \ourMethod's effect on the known object detection is minimal with very scarce data -- however, in these same experiments, the unknown object mAP is significantly higher. At the $100$-shot regime, \ourMethod\ outperforms \baseFS\ both in known and unknown object detection.

\subsection{Zero-Shot Results on the Real-World Object Detection Benchmark}
\label{sup:sec:zs_ood}
It is hard to understand whether or not a particular dataset is out-of-distribution compared to an open-vocabulary object detection model. For example, OWL-ViT utilizes a pre-trained CLIP model. As OpenAI has yet to make its WIT open-source, it is impossible to determine what data it has seen. When studying the models' tokenizer, a clear trend can be seen in the number of words found in the tokenizer (and do not require to be "broken down" to be tokenized). However, in some domains, like the Aerial, we would expect to have a higher \% in tokenizer as they only contain very common items (see Tab.~\ref{table:dataset_details_ood}).

A better indicator is the zero-shot results of models on different benchmarks. Our ``ZS'' baselines (\base, \baseIN, \baseLLM, and \baseGT). These baselines have high performance on traditional open world object detection benchmarks, such as M- and S-OWODB. Meanwhile, these same baselines had a near-zero performance on the proposed real-world object benchmark across most of the application datasets. We can therefore understand that our base model has not frequently seen RWD classes during training.

\subsection{Real-World Object Detection Benchmark - Full Ablation}
\label{sup:sec:full_ablation}
In addition to the ablation results in Tab.~\ref{table:ablation}, we provide the full ablation experiments (with dataset breakdown) in Tab.~\ref{table:full_ablation}. Specifically, each of our contributions seems to be uniformly improving performance. We also show near-zero performance when we remove Adaptation (no longer adjust for the modality gap). As discussed in Sec.~\ref{sec:ablation}, attribute selection sees the most significant improvement overall.

\subsection{Attribute Investigation}
\label{sup:sec:attribute_study}
To study the relationship of the known and unknown classes to the attributes, we collected and averaged the scores of all attributes for each class across the evaluation dataset. We then selected the highest-scoring (on average) attribute per class and visualized these through attribute-class heatmaps in Fig.~\ref{fig:att_study}. Fig~\ref{fig:att_study}, which effectively showcases the attribute prevalence and distinctiveness across the known and unknown classes, where the average scores for each attribute (X-axis) with each class (Y-axis). Across all the datasets, the same attributes frequently had high average scores for more than one class. In some datasets, the attributes were more shared (for example, in the Game, Medical, and Surgery datasets), while others they were more separated (such as the Aerial and Aquatic datasets). We can see that the objects selected for unknown also had high average scores for these same attributes.

\subsection{Additional Qualitative Results}
\label{sup:sec:more_qual}
In this section, we present additional qualitative results. 
In Fig.~\ref{fig:qualitative_good}, we highlight examples from RWD where \ourMethod\ performs favorably while Fig.~\ref{fig:qualitative_bad} shows failure cases. 
 All figures were generated with the $100$-shot L/14 models. 

 \paragraph{\ourMethod\ Favorable Performance.} In Fig.~\ref{fig:qualitative_good},
 \ourMethod\ seems to have consistently better unknown object detection performance. For example, in the Aquatic dataset, \ourMethod\ successfully recognizes the three starfish present on image, while \baseFS\ only detected a single instance. In the Aerial dataset, \ourMethod\ identifies the boat port as a plausible unknown object. In the Game dataset, \ourMethod\ accurately picks up the unknown character, whereas \baseFS\ cannot. 
In the Medical dataset, both \ourMethod\ and \baseFS\ detect similar known object classes. However, \baseFS\ incorrectly identifies the logo as an unknown object, abd misses many bones as possible unknown while \ourMethod\ does not. Meanwhile, it also appeared to frequently perform more favorably in known object detection. For example, in the Aerial dataset, \ourMethod\ identifies most boats correctly, whereas \baseFS\ does not. 
Finally, in the Surgery dataset, while \baseFS\ fails to detect the known Suction Tube, \ourMethod\ accurately detects it.

 \paragraph{\ourMethod\ Failure Cases.} In Fig.~\ref{fig:qualitative_bad}, we highlight images where \ourMethod\ did not perform favorably. In terms of unknown object detection, in the Aquatic dataset, \ourMethod\ fails to identify the second, smaller fish as an unknown object.
In the Aerial dataset, \ourMethod\ failed at identifying any unknown objects, while the \baseFS\ detected one of the unknown objects.
In the Game dataset, Both \ourMethod\ and \baseFS\ cannot identify all unknown characters.
In the Medical dataset, while \ourMethod\ can identify slightly more objects than \baseFS\, it incorrectly identifies them as unknown objects.
In the Medical and Surgical datasets, \ourMethod\ and \baseFS\ failed to detect multiple known objects.

 \section{Limitations}
 \label{sec:limitations}
 In this work, we aim to study open world detection in the era of foundation models.
However, the definition of an unknown becomes fuzzy when we integrate such models as they have been pre-trained on large datasets that can contain any objects. 
Therefore, it is impossible to directly compare our baselines to previous OWD methods such as PROB and OW-DETR. Unlike these methods, our baselines undoubtedly received some supervision during training for each task's `unknown' classes. This exact definition limits the utilization of foundation models on traditional OWD tasks. 

Our definition of an unknown is objects whose labels are not provided by the user, making zero-shot methods not directly applicable. 
We believe this fuzziness should not be an issue from a utility perspective since what is seen and unseen during pre-training is independent of what tasks an end user might care about. 
Empirically, our results showed that \ourMethod\ outperformed many more straightforward methods utilizing foundation models. However, from the perspective of understanding foundation models, future studies that investigate the different levels of unknowns and resolve the fuzzy definition of what is an unknown will bolster the research direction of reliably using foundation models.

{
    \small
    \bibliographystyle{ieeenat_fullname}
    \bibliography{main}
}
\end{document}